\documentclass[lettersize,journal]{IEEEtran}
\usepackage{amsmath,amsfonts,amssymb,amsthm}
\usepackage{algorithm}
\usepackage{array}
\usepackage[caption=false,font=normalsize,labelfont=sf,textfont=sf]{subfig}
\usepackage{textcomp}
\usepackage{stfloats}
\usepackage{url}
\usepackage{verbatim}
\usepackage{graphicx}
\usepackage{algpseudocode}

\usepackage{amsmath}
\usepackage{amssymb}
\usepackage{epstopdf}
\usepackage{mathtools}
\usepackage{booktabs}
\usepackage{caption}
\usepackage{multirow}
\usepackage{stfloats} 
\usepackage{bm}
\usepackage{xcolor}
\usepackage{makecell}

\newtheorem{theorem}{Theorem}
\newtheorem{lemma}{Lemma}

\newtheorem{assumption}{Assumption}

\def\BibTeX{{\rm B\kern-.05em{\sc i\kern-.025em b}\kern-.08em
    T\kern-.1667em\lower.7ex\hbox{E}\kern-.125emX}}
\usepackage{balance}
\begin{document}
\title{Optimizing the Adversarial Perturbation with a Momentum-based Adaptive Matrix}
\author{Wei~Tao, ~Sheng~Long, ~Xin~Liu, ~Wei~Li, ~Qing~Tao
\thanks{This work was supported in part by the NSFC (Grant No. 62576351 and 62076252) and in part by the CPSF (Grant No. 2024M764294). }
\thanks{Corresponding author: Qing Tao (taoqing@gmail.com)}
\thanks{Wei Tao and Sheng Long are with the National University of Defense Technology, Changsha, 410073, China. Wei Tao is also with the Academy of Military Science, Beijing 100091, China. Xin Liu is with the Jiangxi University of Finance and Economics, Nanchang, 330032, China. Wei Li is with the Army Arms University of PLA, Hefei, 230031, China. Qing Tao is with the Hefei Institute of Technology, Hefei, 238076, China.}}


\maketitle

\begin{abstract}
Generating adversarial examples (AEs) can be formulated as an optimization problem. Among various optimization-based attacks, the gradient-based PGD and the momentum-based MI-FGSM have garnered considerable interest. However, all these attacks use the sign function to scale their perturbations, which raises several theoretical concerns from the point of view of optimization. In this paper, we first reveal that PGD is actually a specific reformulation of the projected gradient method using only the current gradient to determine its step-size. Further, we show that when we utilize a conventional adaptive matrix with the accumulated gradients to scale the perturbation, PGD becomes AdaGrad. Motivated by this analysis, we present a novel momentum-based attack AdaMI, in which the perturbation is optimized with an interesting momentum-based adaptive matrix. AdaMI is proved to attain optimal convergence for convex problems, indicating that it addresses the non-convergence issue of MI-FGSM, thereby ensuring stability of the optimization process. The experiments demonstrate that the proposed momentum-based adaptive matrix can serve as a general and effective technique to boost adversarial transferability over the state-of-the-art methods across different networks while maintaining better stability and imperceptibility.
\end{abstract}


\section{Introduction}
\label{sec:intro}
Deep learning has achieved widespread success especially in computer vision \cite{krizhevsky2012imagenet} and natural language procession \cite{Devlin2019BERTPO}. However, evidence has shown that it is challenged by the vulnerability to adversarial attacks \cite{Szegedy2014IntriguingPO}. An adversarial example (AE) is a carefully crafted sample with subtle perturbations that are imperceptible to humans, but can deceive the model to output incorrect results. One interesting phenomenon is that AEs can often transfer from one model to another. The intuition behind this adversarial transferability is that different models learn almost similar classifiers, making it possible to implement attacks without access to target models \cite{Szegedy2014IntriguingPO, huang2023erosion}.
AEs can cause severe threats in security-sensitive applications, such as autonomous driving \cite{zhang2018camou} and intrusion detection \cite{zhang2025explainable}.

Generating AEs can be formulated as an optimization problem of maximizing the loss function with box-constraints \cite{Madry2018TowardsDL, Goodfellow2015ExplainingAH}. So far, many optimization-based methods have been proposed \cite{Carlini2017TowardsET,Yu2023ReliableEO, Gu2023survey, Ge2023Boosting}. Typical gradient-based instances include FGSM \cite{Goodfellow2015ExplainingAH}, I-FGSM \cite{Kurakin2017AdversarialEI}, PGD \cite{Madry2018TowardsDL}, and they are directly motivated by the projected gradient method (PGM). Momentum-based methods, which utilize the accumulation of past gradients, can stabilize the update directions and escape from poor local optimum. To date, two kinds of momentum-based attacks have emerged. One is MI-FGSM \cite{Dong2018BoostingAA}, grounded in Polyak's heavy-ball (HB) method \cite{Polyak1964SomeMO}, and the other is NI-FGSM \cite{Lin2020NesterovAG}, derived from Nesterov's accelerated gradient (NAG) \cite{nesterov27method}. In practice, the momentum-based attacks usually exhibit high transferability while the gradient-based attacks often generate AEs with potential imperceptibility \cite{yang2025A}. To further improve the transferability, various momentum-based attack such as VMI-FGSM \cite{Wang2021EnhancingTT}, EMI-FGSM \cite{Wang2021BoostingTT}, IE-FGSM \cite{Peng2023Boosting}, MIG \cite{Ma2023Transferable}, GRA \cite{Zhu2024GRA}, RAP \cite{Zhu2024RAP}, PGN \cite{Ge2023Boosting} and NCS \cite{Qiu2024Enhancing} have been developed. However, all these optimization-based attacks differ from their counterparts in optimization theory, as they exclusively employ a sign function to scale their perturbations.

Sign function can scales the gradients so that each generated AE easily satisfies the constrains. However, in comparison to the fundamental PGM, I-FGSM can only be regarded as an empirical approach to deal with constrained problems. From an optimization perspective, several theoretical concerns are inevitably raised. First of all, sign-gradient is popularly used only in distributed optimization due to its effectiveness in compressing the gradient to alleviate the communication bottleneck. {\it Why is it so predominantly used in adversarial attacks, and what is the relationship between the gradient-based attacks and their conventional gradient methods?} Secondly, once an optimization algorithm is concerned, its convergence may be one of the most important issues. The convergence analysis of an algorithm heavily depends on its update direction and step-size rule \cite{bertsekas2003convex}. Nevertheless, I-FGSM and PGD empirically use the sign-gradient as its update direction rather than the real gradient. Several simple convex counter-examples have shown that sign-gradient does not converge \cite{Karimireddy2019ErrorFF}. Although it can converge \cite{Bernstein2018signSGDCO}, strong assumptions such as smoothness are required on the objective functions \cite{Karimireddy2019ErrorFF}. {\it Can we have some other way to scale the gradients while keeping their convergence?} Thirdly, evidence has shown that updating direction with the accumulated gradients brings remarkable improvement in stability and transferability \cite{Dong2018BoostingAA, Lin2020NesterovAG}. {\it Can we further use the accumulated gradients to stabilize the whole process of optimization?} Finally, the limitations of PGD, including non-convergence, sensitivity to hyperparameters, budget constraints, and a lack of adaptability due to fixing the step-sizes have been identified \cite{Croce2020ReliableEO, Yuan2024Adaptive}. Besides, as the step-sizes of both I-FGSM and MI-FGSM are decided by the total number of iterations, this strategy precludes the generation of AEs on the fly when we try different numbers of iterations in applications. {\it Can these weaknesses be addressed by employing time-varying step-sizes?} The motivation of this paper is to address these issues by leveraging the adaptive gradient idea from optimization theory while enhancing the adversarial transferability.

Adaptive gradient has demonstrated its effectiveness in mitigating a key limitation of PGM, which is the uniform scaling of each element in the update direction. Pioneering this approach, AdaGrad \cite{Duchi2010AdaptiveSM} calculates the sum of squared past gradients to dynamically adjust the scale of each gradient coordinate with an adaptive diagonal matrix. In essence, AdaGrad can be understood as a second-order gradient technique where the Hessian matrix of the objective function is approximated by a diagonal matrix. This method retains the same convergence rate as PGM while exhibiting a more favorable convergence factor in sparse learning. AdaGrad also has led to the development of various adaptive techniques in deep learning. For example, RMSProp \cite{tieleman2012lecture} utilizes an exponential moving average (EMA) in place of a cumulative sum to effectively forget past gradients. Adam \cite{Kingma2015AdamAM}, currently a leading training algorithm in deep learning, employs the gradient-based adaptive diagonal matrix to scale its step-size and the momentum.

Several interesting adaptive gradient techniques have been introduced to generate AEs. For instance, Ada-FGSM \cite{Shi2020AdaptiveIA} and APAA \cite{Yuan2024Adaptive} were proposed, where the step-size is dynamically optimized as a scalar based on gradient information at each iteration, and these approaches achieve higher success rates. To address the limitations of PGD, Auto-PGD (APGD) was developed \cite{Croce2020ReliableEO}, which adapts the step-size according to the overall budget and the progression of optimization process. In \cite{Zou2022MakingAE} and \cite{LongAAAI}, some variants of Adam were introduced to generate indistinguishable and transferable AEs. Additionally, \cite{Yang2022AdversarialEG} incorporates the Adabelief optimizer \cite{Zhuang2020AdaBeliefOA} to enhance the transferability of AEs.

In this paper, we first analyze the correlation between PGD and PGM. This analysis motivates us to devise novel attacks by scaling the gradient with an adaptive diagonal matrix, as an alternative to the sign function. Our adaptive strategy, in comparison to those in \cite{Shi2020AdaptiveIA, Croce2020ReliableEO, Yuan2024Adaptive}, is element-wise. Unlike the studies in \cite{Zou2022MakingAE, Yang2022AdversarialEG, Shi2020AdaptiveIA, LongAAAI}, the adaptive methods presented here are directly established upon the existing optimization-based attacks while ensuring their convergence. In contrast to the investigations in conventional optimization, we use a new momentum-based adaptive matrix to guarantee imperceptibility of the crafted AEs. It is an interesting adaptive mechanism offers distinct advantages in terms of performance and stability, while may also ‌shedding new light on adaptive momentum algorithm design. The key contributions of this paper can be outlined as follows,
\begin{itemize}
\item We show that PGD is a specific reformulation of the regular PGM, utilizing only the current gradient to determine its step-size. Further, PGD becomes AdaGrad when we replace the sign function with an adaptive diagonal matrix using the accumulated gradients.
\item We present a novel momentum-based attack AdaMI based on MI-FGSM, in which the perturbation is optimized with a momentum-based adaptive matrix instead of the sign function and the conventional gradient-based matrix. We prove that AdaMI attains optimal convergence for general convex problems, indicating that AdaMI overcomes the non-convergence of MI-FGSM and thus can guarantee stability of the whole optimization process. 
\item  The experiments demonstrate that the proposed momentum-based adaptive matrix can serve as a general and effective technique to boost the adversarial transferability across different neural network architectures while maintaining better stability and imperceptibility. Specifically, as far as we know, the derived AdaNCS achieves the best transferability among all the optimization-based attacks.
\end{itemize}

\section{Optimization Problems}
\label{section2}
In this section, we describe the optimization problem for adversarial attacks.

Let $\mathcal{S}=\{(\bm{x}_1, y_1), \dots, (\bm{x}_m, y_m)\}$ be a training set, where $y_i$ is the label of $\bm{x}_i \in \mathbb{R}^{d} $. 
Given a classifier $f_{\bm{\theta}}$ with a predefined $\bm{\theta}$, generating a non-targeted AE $\bm{x}^{adv}$ from a real example $\bm{x}$ can be formulated as a constrained optimization problem \cite{Madry2018TowardsDL, Goodfellow2015ExplainingAH},
\begin{equation}\label{adv-optimization}
\max J(f_{\theta}(\bm{x}^{adv}),y), \ s. t. \  \|\bm{x}^{adv}-\bm{x}\|_p \leq \epsilon,
\end{equation}
where $J(f_{\theta}(\bm{x}),y)$ is the loss function. Obviously, optimization problem (\ref{adv-optimization}) coincides with our intuition, i.e., adversarial attack is to find an example $\bm{x}^{adv}$ that misleads the model prediction (i.e., $ f_{\bm{\theta}}(\bm{x}^{adv})\neq y$) while the $l_p$-norm of the {\bf adversarial perturbation} $\|\bm{x}^{adv}-\bm{x}\|_p$ should be restricted to a threshold $\epsilon$. 

Alternatively, learning adversarial examples can also be described as a {\it regularized} optimization problem \cite{Carlini2017TowardsET}
$$
\min \lambda\|\bm{x}^{adv}-\bm{x}\|_p- J(\bm{x}^{adv},y),
$$
where $\lambda$ is the trade-off parameter. In this paper, we only consider the cross-entropy loss and $p=\infty$. For convenience, $J(f_{\theta}(\bm{x}^{adv}),y)$ will be rewritten as $J(\boldsymbol{x}^{adv}_{t})$.

In optimization community, PGM is a one of most fundamental algorithms for solving constrained problems. To solve (\ref{adv-optimization}), its iteration becomes
\begin{equation}\label{PGM}
\bm{x}^{adv}_{t+1}= P_{\mathcal{B}_{\epsilon}}\left(\bm{x}^{adv}_{t} + \alpha_t \nabla_{\bm{x}} J(\bm{x}^{adv}_{t})\right),
\end{equation}
where $\nabla_{\bm{x}}J(\bm{x})$ is the gradient of $J(\bm{x})$ $w. r. t.$ $\bm{x}$, $\alpha_t>0$ is the time-varying step-size and $P_{\mathcal{B}_{\epsilon}}({\cdot})$ is the projection operator on ${\mathcal{B}_{\epsilon}}=\{\bm{z} \in \mathbb{R}^{d}: \| \bm{z}-\bm{x}\|_\infty \leq \epsilon\}$ \cite{bertsekas2003convex}.

AdaGrad \cite{Duchi2010AdaptiveSM} is a variant of PGM (\ref{PGM}). It takes the form of
\begin{equation}\label{AdaGrad}
\bm{x}^{adv}_{t+1}=P_{\mathcal{B}_{\epsilon}}\left(\bm{x}^{adv}_{t}+\alpha_t V_{t}^{-\frac{1}{2}} \nabla_{\bm{x}} J(\bm{x}^{adv}_{t})\right),
\end{equation}
where $V_{t}$ is a $d \times d$ diagonal matrix and each
\begin{equation}\label{arithmetic average}
v_{t,i}=\frac{\sum_{j=1}^{t}{\left( \nabla_{\bm{x}} J(\bm{x}^{adv}_{j})\right)_{i}^{2}}}{t}
\end{equation}
is the arithmetic average of the square of the $i$-th elements of the past gradients. Here, $v_{t,i}$ denotes the $i$-th element in the diagonal of $V_t$. In this paper, we refer to $V_t$ (\ref{arithmetic average}) as {\bf gradient-based adaptive matrix}. Obviously, the seldom-updated weights are updated with a larger step-size than the frequently-updated weights. Such an adaptive mechanism is well-suited for sparse learning problems \cite{wang2019sadam}.

HB \cite{Polyak1964SomeMO} is popularly used in deep learning \cite{Ruder2016AnOO}. For an unconstrained optimization problem, HB can be written as a general two-steps algorithm,
\begin{equation}\label{two-step}
\left  \{
\begin{array}{l}
\bm{g}_{t+1}=\mu_t \ \bm{g}_t  + \ \nabla_{\bm{x}} J(\bm{x}_{t}^{adv})\\
\bm{x}_{t+1}^{adv}=\bm{x}_{t}^{adv} + \alpha_t \bm{g}_{t+1}
\end{array},
\right.
\end{equation}
where $\mu_t>0$ is the decay factor. Under mild conditions, it can yield an accelerated convergence for both general convex and smooth strongly-convex functions \cite{Polyak1964SomeMO, Tao2021TheRO}.

\section{Optimization-based Attacks}
\label{optimization-based attacks}
In this section, we provide a brief overview of several typical optimization-based attacks.

FGSM \cite{Goodfellow2015ExplainingAH} is a basic gradient-based attack. It has only one-step update, i.e.,
\begin{equation}\label{FGSM}
\bm{x}^{adv}=\bm{x}+\epsilon \ \mathrm{sign}(\nabla_{\bm{x}} J(\bm{x})),
\end{equation}
where $\mathrm{sign}(\cdot)$ is the sign function. From (\ref{FGSM}), it is easy to know $\|\bm{x}^{adv}-\bm{x}\|_\infty \leq \epsilon$.

I-FGSM \cite{Kurakin2017AdversarialEI} is a FGSM with multiple iterative steps, i.e.,
\begin{equation}\label{I-FGSM}
\bm{x}_{t+1}^{adv}= \bm{x}_{t}^{adv} + \alpha \ \mathrm{sign}(\nabla_{\bm{x}} J(\bm{x}_{t}^{adv})),
\end{equation}
where $\bm{x}^{adv}_0 = \bm{x}$. Unlike FGSM, the step-size $\alpha$ is set to $\epsilon/T$ so that $\|\bm{x}_{t}^{adv}-\bm{x}\|_\infty \leq \epsilon,\  0 \leq \forall t\leq T$, where $T$ is the total number of iterations. Obviously, FGSM and I-FGSM are motivated by PGM (\ref{PGM}), and the only difference is that FGSM and I-FGSM use the sign function to scale 
$\nabla_{\bm{x}} J(\bm{x}_{t}^{adv})$.

PGD \cite{Madry2018TowardsDL} is a clipped I-FGSM but starting from a random perturbation around the natural example, i.e.,
\begin{equation}\label{PGD}
\bm{x}_{t+1}^{adv} = \mathrm{Clip}_{\bm{x}}^\epsilon \left(\bm{x}_{t}^{adv} + \alpha \ \mathrm{sign} (\nabla_{\bm{x}} J(\bm{x}_{t}^{adv}))\right),
\end{equation}
For an image $\bm{x}=(x_1,x_2,x_3)$ which is typically 3-D tensor, its clipping is \cite{Kurakin2017AdversarialEI}
\begin{equation}\label{clip}
\begin{aligned}
& \mathrm{Clip}_{\bm{x}}^\epsilon \left(\bm{x}(x_1,x_2,x_3)\right)\\
& =\min \{ 255, \bm{x}(x_1,x_2,x_3)+\epsilon,\\
& \max\{0, \bm{x}(x_1,x_2,x_3)-\epsilon, \bm{x}(x_1,x_2,x_3)\}\}.
\end{aligned}
\end{equation}
Obviously, the clipping (\ref{clip}) ensures that $\|\bm{x}^{adv}-\bm{x}\|_\infty \leq \epsilon$. Unlike I-FGSM, the clipping is necessary since a random perturbation is used and the step-size is not restricted.

The momentum-based attack MI-FGSM \cite{Dong2018BoostingAA} integrates HB momentum \cite{Polyak1964SomeMO} into the iteration of I-FGSM. The update procedure of MI-FGSM is
\begin{equation}\label{MI}
\left  \{
\begin{array}{l}
\bm{g}_{t+1}=\mu \ \bm{g}_t  +  \displaystyle\frac{\nabla_{\bm{x}} J(\bm{x}_{t}^{adv})}{\|\nabla_{\bm{x}} J(\bm{x}_{t}^{adv})\|_1}\\
\bm{x}_{t+1}^{adv}=\mathrm{Clip}_{ \bm{x}}^\epsilon \left(\bm{x}_{t}^{adv} + \alpha \ \mathrm{sign}(\bm{g}_{t+1})\right)
\end{array},
\right.
\end{equation}
where $\alpha=\epsilon/T$.

Similarly, the NAG momentum \cite{nesterov27method} can also be integrated into I-FGSM. The iterative version of momemtum-based attack NI-FGSM \cite{Lin2020NesterovAG} is formulated as
\begin{equation}\label{NI}
\left  \{
\begin{array}{l}
\bm{x}^{nes}_t= \bm{x}^{adv}_t  +\alpha  \bm{g}_t \\
\bm{g}_{t+1}=\mu \ \bm{g}_t  + \displaystyle\frac{\nabla_{\bm{x}} J(\bm{x}_{t}^{nes})}{\|\nabla_{\bm{x}} J(\bm{x}_{t}^{nes})\|_1} \\
\bm{x}_{t+1}^{adv}=\mathrm{Clip}_{\bm{x}}^\epsilon  \left(\bm{x}_{t}^{adv} + \alpha \ \mathrm{sign}(\bm{g}_{t+1})\right)
\end{array}.
\right.
\end{equation}
Compared with I-FGSM (\ref{I-FGSM}) and PGD (\ref{PGD}), the current gradient is replaced with the momentum. The momentum can stabilize their iterative direction $ \mathrm{sign}(\bm{g}_{t+1})$ and the adversarial transferability is then boosted \cite{Dong2018BoostingAA}.

\section{The Connection between PGD and PGM}
In this section, we analyze the connection between PGD (\ref{PGD}) and PGM (\ref{PGM}).

We first indicate that $\mathrm{Clip}_{\bm{x}}^\epsilon(\bm{x}^{adv})$ in (\ref{clip}) is a projection of $\bm{x}^{adv}$ on a specific $\mathbf{Q}$. For example, when an image is described as 3-D tensor, we usually set
$$\mathbf{Q}=\{\bm{z}: \| \bm{z}-\bm{x}\|_\infty \leq \epsilon\}\bigcap [0,255]^3.$$
It is easy to find
$$P_{\mathbf{Q}}(\bm{x}^{adv})= \mathrm{Clip}_{\bm{x}}^\epsilon(\bm{x}^{adv}).$$

In contrast to PGM (\ref{PGM}), PGD (\ref{PGD}) normalize each current gradient with a sign-function. In general cases, we know $\mathrm{sign}\left(\nabla_{\bm{x}}J(\bm{x}^{adv}_{t})\right) \neq \nabla_{\bm{x}}J(\bm{x}^{adv}_{t})$, i.e., PGM (\ref{PGM}) and PGD (\ref{PGD}) have different update magnitudes. Fortunately, if we reset the diagonal matrix $V_{t}$ in (\ref{arithmetic average}) as
\begin{equation}\label{not-historical}
v_{t,i}= \left(\nabla_{\bm{x}} J(\bm{x}^{adv}_{t})\right)_{i}^2,
\end{equation}
we have
$$\mathrm{sign}\left(\nabla_{\bm{x}}J(\bm{x}^{adv}_{t})\right)= V_{t}^{-\frac{1}{2}} \left(\nabla_{\bm{x}}J(\bm{x}^{adv}_{t})\right),$$
which reveals that PGD (\ref{PGD}) is in fact a specific PGM (\ref{PGM}) but using $\nabla_{\bm{x}}J(\bm{x}^{adv}_{t})$ to determine its step-size. However, even for convex objective functions, the step-size (\ref{not-historical}) will cause non-convergence of PGM (\ref{PGM}) \cite{Karimireddy2019ErrorFF}. Thus, we don't know if $J(\bm{x}_{t}^{adv})$ is convergent to $ \max J(\bm{x}^{adv})$ on $  \|\bm{x}^{adv}-\bm{x}\|_p \leq \epsilon$. Further, PGD may generalize poorly compared with SGD when solving machine learning problems \cite{Karimireddy2019ErrorFF}.

The aforementioned analysis implies that PGD can be extended to adaptive cases by scaling its perturbation with the accumulated gradients. This kind of extension is also encouraged by the success of the momentum-based attacks. Specifically, we can replace its sign function with a gradient-based adaptive matrix (\ref{arithmetic average}). Interestingly, it can be observed that the derived adaptive PGD exactly coincides with AdaGrad (\ref{AdaGrad}). The detailed steps of AdaGrad for adversarial attacks are shown in Algorithm  \ref{alg:AdaGrad}.

\begin{algorithm}[tb]
\caption{AdaGrad}
\label{alg:AdaGrad}
\begin{algorithmic}
   \State {\bfseries Input:} The perturbation size $\epsilon$, step-size parameter $\alpha>0$ and constant $\delta > 0$.
   \State {\bfseries Initialize:} $\bm{x}^{adv}_0=\bm{x}$ and $V_0 = \mathbf{0}_{d\times d}$.
   \Repeat
   \State Update $V_t$ by Eq. (\ref{arithmetic average}).
   \State $\hat{V_t} = V_{t}+\delta I_d$.
   \State $\bm{x}_{t+1}^{adv}= P_{\mathbf{Q}} \left(\bm{x}_{t}^{adv} + \displaystyle\frac{\alpha }{\sqrt{t}} \hat{V_t}^{-\frac{1}{2}} \nabla_{\bm{x}}J(\bm{x}^{adv}_{t})\right).$ 
   \Until {$t = T$}
   \State {\bfseries Return:} $\bm{x}^{adv}_T$.
\end{algorithmic}
\end{algorithm}

In Algorithm \ref{alg:AdaGrad}, a vanishing factor $\delta I$ is added to the diagonal of $V_t$ and get $\hat{V_t}$. Such an operation is a standard technique to avoid too large steps caused by zero or small gradients in the beginning iterations \cite{Ruder2016AnOO}. AdaGrad enjoys an $O(\sum_{i=1}^{d}\| \nabla_{\bm{x}}J(\bm{x}^{adv}_{t})_{1:t,i}\|)$ regret for general convex functions, where $g_{t,i}$ denotes the $i$-th element of $\boldsymbol{g}_t$ and $\boldsymbol{g}_{1:T, i}=[g_{1,i}, \cdots, g_{T,i}]$ is the vector obtained by concatenating the $i$-th element of the sequence $\{\boldsymbol{g}_t\}^T_{t=1}$. This regret is $O(d\sqrt t)$ in the worst case and becomes tighter when gradients are sparse \cite{Duchi2010AdaptiveSM}. In strongly convex cases, the regret bound can be further improved \cite{wang2019sadam}.

Our connection analysis is also applicable to I-FGSM. As the element-wise adaptive step-size no longer guarantees that $\|\bm{x}^{adv}-\bm{x}\|_\infty \leq \epsilon$, the projection operator $P_{\mathbf{Q}}$ must be used.

Based upon the aforementioned analysis, we can also get several new variants of adaptive gradient-based attacks by using different kinds of $V_t$. For example, 
we can use EMA strategy and $l_1$-norm,
\begin{equation}\label{l1-adaptive}
v_{t,i}=\beta v_{t-1,i}+ (1-\beta) |\left( \nabla_{\bm{x}} J(\bm{x}^{adv}_{t}) \right)_{i}|,
\end{equation}
where $1>\beta>0$. Adaptive attack (\ref{l1-adaptive}) is also an extension of PGD but with guaranteed convergence.

\section{The Proposed AdaMI}
\label{sec:AdaMI-G}

A natural adaptive variant of MI-FGSM can be easily derived by directly employing the gradient-based adaptive matrix (\ref{arithmetic average}) to adjust the step size. We denote such an adaptive variant as {\bf AdaMI-G}. The detailed steps of AdaMI-G are described in Algorithm \ref{alg:AdaMI-G}.

\begin{algorithm}[htb]
\caption{AdaMI-G}
\label{alg:AdaMI-G}
\begin{algorithmic}
   \State {\bfseries Input:} The perturbation size $\epsilon$, step-size parameter $\alpha_t>0$, EMA parameter $1\geq\beta\geq0$, constant $\delta > 0$ and momentum parameter $\mu_t>0$.
   \State {\bfseries Initialize:} $\bm{g}_0=0$, $\bm{x}^{adv}_0=\bm{x}$ and $V_0 = \mathbf{0}_{d\times d}$.
   \Repeat
   \State $\bm{g}_{t+1}=\mu_t \ \bm{g}_t  + \displaystyle\frac{\nabla_{\bm{x}} J(\bm{x}_{t}^{adv})}{\|\nabla_{\bm{x}} J(\bm{x}_{t}^{adv})\|_1}$.\\
   \State Update the diagonal matrix $V_{t+1}$ via $$v_{t+1,i}=\beta v_{t,i}+(1-\beta)\left( \nabla_{\bm{x}} J(\bm{x}_{t}^{adv})\right)_{i}^{2}.$$
   \State $\hat{V}_{t+1} = V_{t+1} +\delta I_d$.
   \State $  \bm{x}_{t+1}^{adv}= P_{\mathbf{Q}} \left(\bm{x}_{t}^{adv} +\alpha_t \hat{V}_{t+1}^{-\frac{1}{2}} \bm{g}_{t+1}\right).$
   \Until {$t = T$}
   \State {\bfseries Return:} $\bm{x}^{adv}_T$.
\end{algorithmic}
\end{algorithm}

Nevertheless, drawing inspiration from the connection between PGD and AdaGrad, we will present a new adaptive algorithm {\bf AdaMI} based on AdaMI-G, which is one of the contributions in this paper. The detailed steps of AdaMI are described in Algorithm \ref{alg:AdaMI}.

\begin{algorithm}[htb]
\caption{AdaMI}
\label{alg:AdaMI}
\begin{algorithmic}
   \State {\bfseries Input:} The perturbation size $\epsilon$, step-size parameter $\alpha_t>0$, EMA parameter $1\geq\beta\geq0$, constant $\delta > 0$ and momentum parameter $\mu_t>0$.
   \State {\bfseries Initialize:} $\bm{g}_0=0$, $\bm{x}^{adv}_0=\bm{x}$ and $V_0 = \mathbf{0}_{d\times d}$.
   \Repeat
   \State $\bm{g}_{t+1}=\mu_t \ \bm{g}_t  + \displaystyle\frac{\nabla_{\bm{x}} J(\bm{x}_{t}^{adv})}{\|\nabla_{\bm{x}} J(\bm{x}_{t}^{adv})\|_1}$.\\
   \State Update the diagonal matrix $V_{t+1}$ via $$v_{t+1,i}=\beta v_{t,i}+(1-\beta)\left( \bm{g}_{t+1}\right)_{i}^{2}.$$
   \State $\hat{V}_{t+1} = V_{t+1} +\delta I_d$.
   \State $  \bm{x}_{t+1}^{adv}= P_{\mathbf{Q}} \left(\bm{x}_{t}^{adv} +\alpha_t \hat{V}_{t+1}^{-\frac{1}{2}} \bm{g}_{t+1}\right).$
   \Until {$t = T$}
   \State {\bfseries Return:} $\bm{x}^{adv}_T$.
\end{algorithmic}
\end{algorithm}

It should be noted that AdaMI-G with gradient-based adaptive matrix is a direct variant of MI-FGSM, while AdaMI employs momentum-based adaptive matrix. In fact, the update of each element in the diagonal adaptive matrix of AdaMI-G is
\begin{equation}
v_{t+1,i}=\beta v_{t,i}+(1-\beta)\left( \nabla_{\bm{x}} J(\bm{x}_{t}^{adv})\right)_{i}^{2}.
\end{equation}
Obviously, such an update is gradient-based. This is different from the momentum-based update in AdaMI, i.e.,
\begin{equation}
v_{t+1,i}=\beta v_{t,i}+(1-\beta)\left( \bm{g}_{t+1}\right)_{i}^{2}.
\end{equation}
The primary novelty of AdaMI lies in its computation of the adaptive matrix $V_t$.
Specifically, each element $v_{t,i}$ is obtained as the EMA of the squares of the corresponding elements from past momentum vectors. This approach fundamentally differs from Adam-type algorithms, which compute the EMA of past gradients instead of past momentums \cite{reddi2019convergence,Ruder2016AnOO,wang2019sadam,Tao2021TheRO}. To the best of our knowledge, this may be a new design paradigm for introducing adaptivity into momentum-based methods, even beyond adversarial attacks to the broader context of optimization. In this paper, we refer to this $V_t$ as {\bf momentum-based adaptive matrix}.

The utilization of a momentum-based adaptive matrix offers several significant benefits in optimizing the adversarial perturbation. First of all, it ensures that the connection between MI-FGSM and AdaMI is completely similar to the correlation between PGD and AdaGrad. So, the main difference between MI-FGSM and AdaMI lies in the way of scaling the perturbation $\bm{g}_{t+1}$. The former uses the sign function while the latter employs an adaptive matrix. Secondly, it guarantees that AdaMI can effectively keep the transferability of MI-FGSM, as we only use the EMA of past momentums to replace the current momentum in MI-FGSM. Thirdly, AdaMI will become MI-FGSM when $\beta=0$, i.e., the momentum-based adaptive matrix matches the perturbation of AdaMI better than the gradient-based matrix and AdaMI can be expected to generate small perturbations and then maintain imperceptibility of the crafted AEs like gradient-based attacks. Finally, the convergence is not affected when we replace the gradient-based adaptive matrix with the momentum-based one, i.e., if we can prove convergence of a specific AdaMI without using the adaptive matrix, the convergence of AdaMI will be derived without additional difficulties. In fact, the proof is similar to that in conventional adaptive scenarios. The detailed proof will be given in Appendix. The experimental comparison between AdaMI-G and AdaMI is provided in Experiments.


To analyze the convergence of AdaMI, several assumptions should be made.
\begin{assumption}
\label{assum}
Assume that the objective function $J(\boldsymbol{x}^{adv})$ is concave. Suppose there exists a number $D_1 > 0$ such that
\begin{equation*}
  \| \boldsymbol{x}_{1}-\boldsymbol{x}_{2}\| \leq D_1,\ \ \forall \boldsymbol{x}_{1}, \boldsymbol{x}_{2} \in \mathbf{Q},
\end{equation*}
and there exists a number $G > 0$ such that
\begin{equation*}
  \| \nabla_{\boldsymbol{x}}J(\boldsymbol{x}^{adv})\| \leq \| \nabla_{\boldsymbol{x}}J(\boldsymbol{x}^{adv})\|_{1} \leq G, \forall \boldsymbol{x}^{adv} \in \mathbf{Q}.
\end{equation*}
\label{ass:xfinite}
\end{assumption}

\begin{theorem}
\label{maintheorem1} Suppose $\boldsymbol{x}^{\ast} $ is a solution of problem (\ref{adv-optimization}). Assume $1>\lambda>0$, $\mu>0$ and $\alpha>0$. Let
$\mu_t=\mu \lambda^{t-1}$ and $ \alpha_t= \displaystyle\frac{\alpha }{\sqrt{t}}.$ Let $\{ \boldsymbol{x}^{adv}_t\}_{t=1}^{\infty} $  and $\{\boldsymbol{g}_t\}_{t=1}^{\infty}$ be generated by AdaMI. Then we have
  $$J(\boldsymbol{x}^\ast)-J(\boldsymbol{\bar{x}}_T^{adv}) \leq O(\frac{\sum_{i=1}^{d}\|\boldsymbol{g}_{1:T,i}\|}{T}),$$
where $\boldsymbol{\bar{x}}_T^{adv} = \frac{1}{T}\sum_{t=1}^{T}\boldsymbol{x}^{adv}_t$.
\end{theorem}

To understand the importance of our theoretical analysis in adversarial attacks, we give some remarks.
\begin{itemize}
\item According to Theorem \ref{maintheorem1}, AdaMI attains optimal averaging convergence for general convex problems. As far as we know, it may be the first momentum-based attack guaranteeing convergence for convex problems. 
\item In contrast to I-FGSM and MI-FGSM, AdaMI uses a time-varying step-size $\alpha_t$, which does not need to know the performed number of iterates in advance. Thus the AEs can be generated on the fly, and this property is highly desirable when we want to run a method using different numbers of iterations.
\item When an AE can be successfully generated from a correctly classified sample $\boldsymbol{x}$, we have $J(\boldsymbol{x}^{adv}) > J(\boldsymbol{x})$. Intuitively, we can assume that the objective function is locally concave. Since the constrained domain is generally small due to the human-imperceptibility constraint, we can further assume the objective function to be concave without loss of generality.
 \item Like that in RMSProp \cite{mukkamala2017variants} and Adam \cite{Kingma2015AdamAM}, we think that our analysis here is not only interesting for overcoming the non-convergence of sign-gradient in the convex case but can give valuable hints how the parameters of AdaMI should be chosen in adversarial attacks.
\end{itemize}

\section{Experiments}
\label{sec:Experiments}

In this section, we conduct experiments to to show the benefits brought by scaling the perturbations with adaptive matrices. Typically, we only focus on comparing our derived adaptive attacks with some recent optimization-based attacks. We keep the standard hyperparameters according to the original algorithm. The success rate of each attack denotes its misclassification rate with AEs as inputs. A popular proxy measure of imperceptibility is the $L_p$-norm of the perturbation. It offers a good trade-off between simplicity, mathematical tractability, and practicality in applications \cite{Chawin2024}. In this paper, we employ the Average $L_{\infty}$ Distortion (ALD$_p$) \cite{Lin2020NesterovAG} and Fréchet Inception Distance (FID) \cite{Martin2017FID} as the indicators of imperceptibility of the crafted AEs. All the experiments are run on a single NVIDIA A30 24GB PCIe GPU.

Different transfer-based attacks exhibit similar performance on CIFAR-10, while some state-of-the-art methods can improve the transferability of AEs on ImageNet \cite{Dong2020BenchmarkingAR}. One potential reason is that the input dimension of the models on ImageNet is much higher than that on CIFAR-10, and the AEs generated by some algorithms may easily overfit the surrogate model. So, we only focus on ImageNet in our experiments. We will use the same dataset as that in \cite{Dong2018BoostingAA, Lin2020NesterovAG}, i.e., we randomly sample 1,000 images from ImageNet validate set  \cite{Russakovsky2015ImageNetLS}, in which each image is from one category and can be correctly classified by the adopted models. Note that these images are resized to $224\times 224\times 3$.
We choose models from both branches of CNNs and ViTs for the black-box attacks, including Inception-v3 (Inc-v3), ResNet-34 (Res-34), VGG-16, Inception-ResNet-v2 (IncRes-v2), MobileNet-v2 (Mob-v2) in CNN branch; ViT-Small (ViT-S), ConViT-Base (ConViT-B), Visformer-Small (Visformer-S), Swin-Tiny-Patch4 (Swin-T) in ViT branch \cite{Alexey2021Vit,Liu2021swin}. Following \cite{Li2023Improving}, we also collect robust 
EfficientNet-B0 (Efficient-B0$_{adv}$), EfficientNet-B1 (Efficient-B1$_{adv}$), Inception-ResNet-v2 (IncRes-v2$_{ens}$),  Inception-v3 (Inc-v3$_{adv}$) from the timm repository\footnote{{\url{https://github.com/rwightman/pytorch-image-models}}}. 

The parameters in problem (\ref{adv-optimization}) are fixed. We set $\epsilon={8}/{255}$ and $T = 10$. As usual, $\delta=10^{-20}$ and $\lambda = 0.999$ \cite{wang2019sadam, Tao2021TheRO, reddi2019convergence}. For convenience, we select $\alpha={\epsilon}/{10}$ in all the concerned algorithms. One important adjustable parameter in momentum-based adaptive attacks is the EMA parameter $\beta$, and it is selected by simple grid search while considering both transferability and imperceptibility.

\subsection{Comparing with Baseline Attacks}
\label{sec:4.2}

In this subsection, we compare AdaGrad with some typical gradient-based attacks including I-FGSM, PGD, APGD, AutoAttack\footnote{{\url{https://github.com/fra31/auto-attack}}} \cite{Croce2020ReliableEO}, and momentum-based attacks including MI-FGSM \cite{Dong2018BoostingAA}, PI-FGSM \cite{Gao2020PI}, RAP \cite{Zhu2024RAP}, GRA \cite{Zhu2024GRA}, ANDA \cite{Fang2024ANDA}, PGN \cite{Ge2023Boosting}, NCS \cite{Qiu2024Enhancing} on different kinds of models. Empirical results confirm that NCS currently achieves highly competitive performance among gradient-based and momentum-based attack methods including VMI-FGSM and RAP\footnote{{\url{https://github.com/Trustworthy-AI-Group/TransferAttack}}}. Based on these findings and considering the central role of NCS as a state-of-the-art approach, we focused our subsequent comparative experiments primarily on MI-FGSM and NCS. For simplicity, the AEs are crafted for Res34. The success rates of attacks are reported in Tab.\ref{tab:gradient}.

\begin{table*}[htbp]\small
\captionsetup{font={normalsize}}
\caption{\normalsize{Transferability comparisons across different network architectures. The imperceptibility is measured by FID. AEs are crafted for Res34. The baseline attacks are divided into two categories: gradient-based attacks and momentum-based attacks. The best results are marked in bold.}}
\label{tab:gradient}
\centering
\begin{tabular}{c|cccccccc|c}%
\toprule 
Attack &Res34& Inc-v3& Dense-121 & IncRes-v2& ViT-S&Swin-T & Efficient-B0$_{adv}$&IncRes-v2$_{ens}$  &FID\\	
\midrule 
PGD & 98.7$^*$&20.5&20.6&9.8& 8.8& 12.2& 10.0& 8.8 &12.985\\
I-FGSM & 99.6$^*$&20.6&24.9&10.3&  8.6& 13.5& 10.3& 8.8 &11.724\\
APGD & 99.6$^*$& 20.8& 22.2& 10.1& 8.7& 11.5& 10.6& 8.2 &10.519\\
AutoAttack & 99.6$^*$& 20.7& 21.5& 10.0& 8.7& 10.7& 10.4& 7.9 &\bf{10.125}\\
AdaGrad& 99.8$^*$& 21.4& 28.4& 10.4& 8.9& 14.4& 11.2& 9.2 &13.469\\ \hline
MI& 100.0$^*$& 36.8& 47.2& 23.4& 17.4& 26.0& 22.9& 15.1 &30.262\\
AdaMI-G& 100.0$^*$& 37.8 & 49.7 & 21.8 & 17.3 & 24.2 & 22.4 & 15.8 & 37.304\\
AdaMI (Ours)& 100.0$^*$& 38.4& 51.5& 24.6& 19.1& 28.0& 25.1& 15.5 &34.599\\
 RAP&100.0$^*$ & 50.3 & 67.4 & 32.4 & 24.7 & 38.5 & 35.4 & 20.4 & 68.034\\
 PGN& 100.0$^*$ & 59.5& 77.3 & 46.7 & 30.8 & 44.8& 47.4 & 32.5 &56.380\\
 ANDA& 100.0$^*$ & 62.4 & 81.9 & 53.8 &31.8& 44.1& 45.9&29.9&78.817\\
 NCS& 100.0$^*$& 62.8 &80.2 &51.5 &35.6 &49.5&52.2&34.9 &60.695\\
AdaNCS (Ours)& 100.0$^*$ & \bf{64.5} & \bf{83.2} & \bf{53.9} & \bf{37.4} & \bf{52.2} & \bf{54.3} & \bf{36.5} &63.471\\ \hline
\end{tabular}
\end{table*}

It can be observed that AdaGrad consistently outperforms other gradient-based attacks on almost all the models. Note that AdaGrad only slightly modifies I-FGSM and PGD by replacing its current gradient with the accumulated gradients to determine its step-size. However, this simple operation can remarkably boosts the transferability. It is worth indicating that the actual perturbation of AdaGrad maintains the same level of I-FGSM and is even better than PGD. Such a fact clearly illustrates our motivation in this paper, i.e., scaling the perturbation with an adaptive matrix is an effective technique to boost the transferability for gradient-based attacks while maintaining better imperceptibility. Meanwhile, our AdaMI and AdaNCS consistently outperforms MI-FGSM and NCS respectively on all the models in terms of the average attack rates. Specifically, AdaNCS achieves the best transferability among all the optimization-based attacks.



\subsection{Integrating into Momentum-based Attacks}
\label{comparsion-M1}

The momentum-based adaptive technique can be readily combined with other momentum-based attacks. In this subsection, we integrate the proposed adaptive technique into some state-of-the-art momentum-based attacks including  MI-FGSM, NI-FGSM, VMI-FGSM \cite{Wang2021EnhancingTT}, EMI-FGSM \cite{Wang2021BoostingTT}, IE-FGSM \cite{Peng2023Boosting}, MIG \cite{Ma2023Transferable}, PGN \cite{Ge2023Boosting} and NCS \cite{Qiu2024Enhancing}.

The success rates of attacks are reported in Tab.\ref{tab:mom}. Generally speaking, our adaptive momentum-based attacks achieve a consistently higher attack rate than their vanilla ones. Specifically, our AdaMI and AdaNCS consistently outperforms MI-FGSM and NCS respectively on all the models in terms of the average rates. These experimental results illustrate that generating adaptive perturbations is a general and effective technique to boost the transferability of momentum-based attacks. As empirical evaluations in \cite{Qiu2024Enhancing} have shown that NCS can significantly improve transferability of available optimization-based attacks, to the best of our knowledge, we can say that our AdaNCS achieves so far the best transferability among all the optimization-based attacks. More experimental results can be found in Tab.\ref{tab:momentum_all}. Therefore, generating momentum-based adaptive perturbations can be regarded as a general technique to boost the adversarial transferability.

\begin{table*}[htbp]\small
\captionsetup{font={normalsize}}
\caption{\normalsize{Transferability comparisons between momentum-based attack and Ada-attack. AEs are crafted for Res34. The imperceptibility is measured by  FID. The best results are marked in bold.}}
\label{tab:mom}
\centering
\begin{tabular}{c|cccccccc|c}%
\toprule 
Attack & Res34& Inc-v3& VGG16& Mob-v2& ViT-S& ConViT-B& Visformer-S& Swin-T &FID\\
\midrule 
MI & 100.0$^*$& 36.8& 47.2& 45.8& 17.4& 12.3& 19.0& 25.9 &\bf{30.262}\\
AdaMI (Ours)& 100.0$^*$& 38.4& 51.5& 49.2& 18.4& 13.3& 20.0& 28.0&34.599\\	 
\hline  
NI & 100.0$^*$&38.1&48.6&46.1&17.8&12.8&19.6&27.4 &31.568\\
AdaNI (Ours)& 100.0$^*$&39.6&51.6&50.7& 19.2&13.1&20.6&28.2 &35.254\\
\hline 
VMI& 100.0$^*$& 48.4& 61.1& 57.0& 25.0& 18.0& 28.1& 36.4 &40.489\\
AdaVMI (Ours)& 100.0$^*$& 49.0& 61.4& 57.4& 25.6& 18.1& 28.5& 37.1 &41.896\\  
\hline 
IE&100.0$^*$&40.2&52.6&50.3&19.6&13.3&20.7&29.1 &34.600\\
AdaIE (Ours)& 100.0$^*$& 41.1& 53.5& 50.7& 19.6& 13.2& 20.9& 28.9 &34.910\\  
\hline 
EMI & 100.0$^*$& 47.4& 62.5& 58.5& 22.1& 16.7& 24.5& 35.2 &43.352\\
AdaEMI (Ours)& 100.0$^*$&47.9&62.8&59.7&22.7&16.7&26.2&35.8&43.660\\ 
\hline 
PGN & 100.0$^*$& 59.5& 71.7& 66.1& 32.3& 25.0& 36.7& 44.8 &56.380\\
AdaPGN (Ours)& 100.0$^*$& 59.9& 72.0& 66.9& 32.5& 25.5& 37.6& 45.4 &56.186\\ 
\hline  
NCS & 100.0$^*$ & 62.8& 75.6& 70.4& 35.5& 29.2& 42.8& 49.5&60.695 \\
AdaNCS (Ours)& 100.0$^*$ & \bf{64.5}& \bf{79.0}& \bf{73.5}& \bf{36.5}& \bf{30.5}& \bf{43.8}& \bf{52.2}&63.471\\ \hline
\end{tabular}
\end{table*}

\begin{figure*}[htbp]\small
\centering
\begin{minipage}{0.4\linewidth}
\centering
\includegraphics[width=\textwidth]{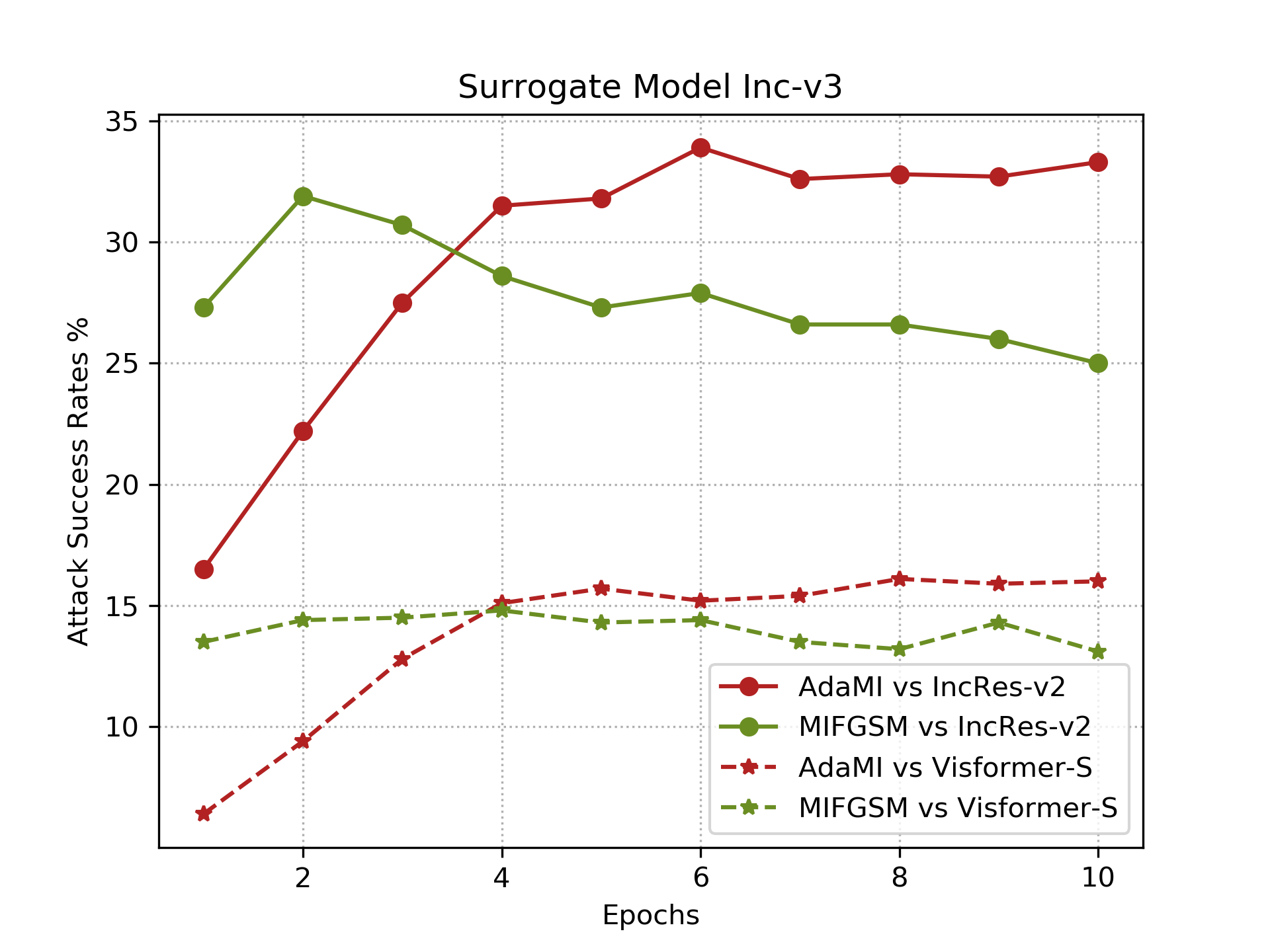}
\end{minipage}
\begin{minipage}{0.4\linewidth}
\centering
\includegraphics[width=\textwidth]{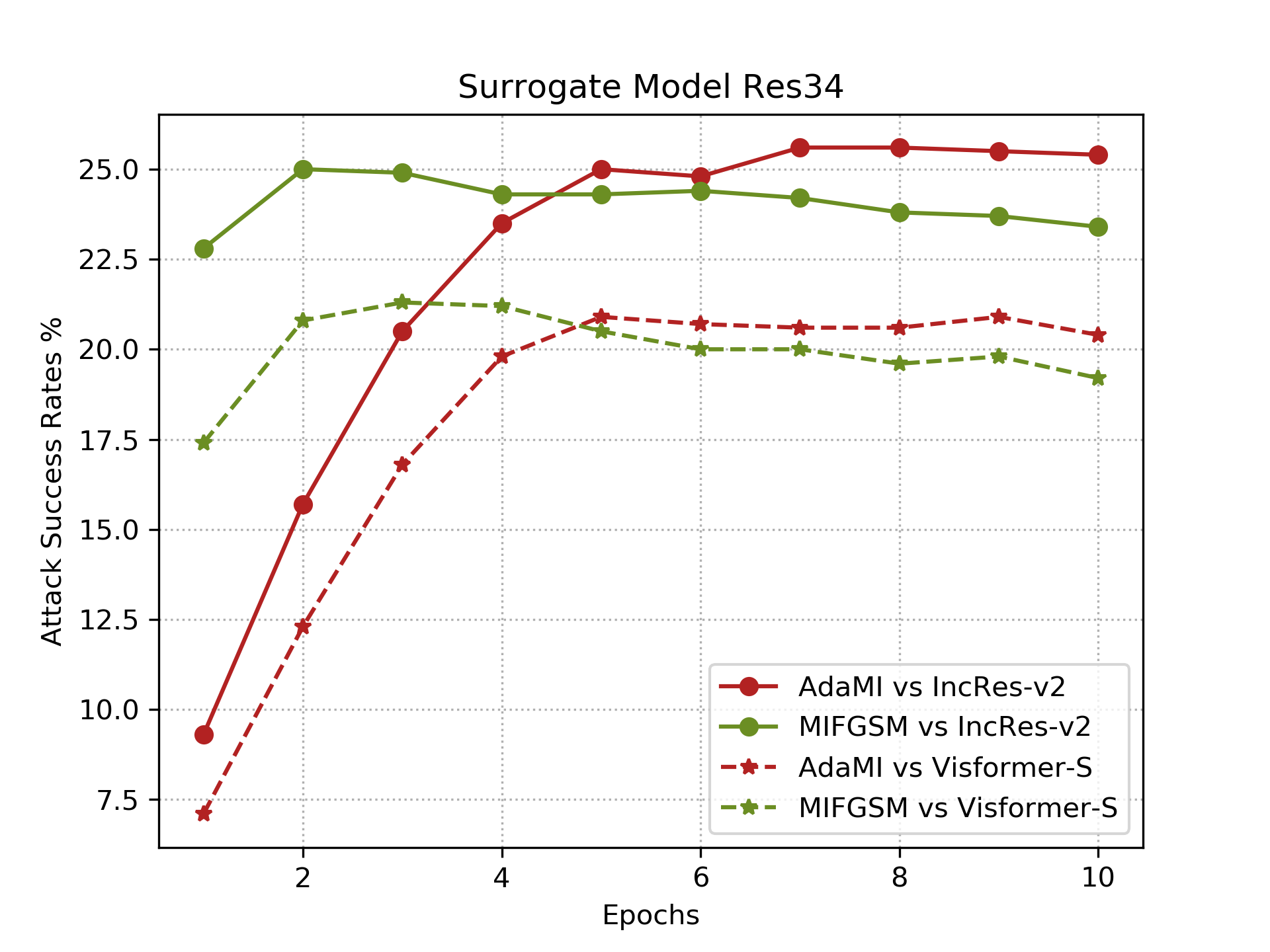}
\end{minipage}\\
\begin{minipage}{0.4\linewidth}
\centering
\includegraphics[width=\textwidth]{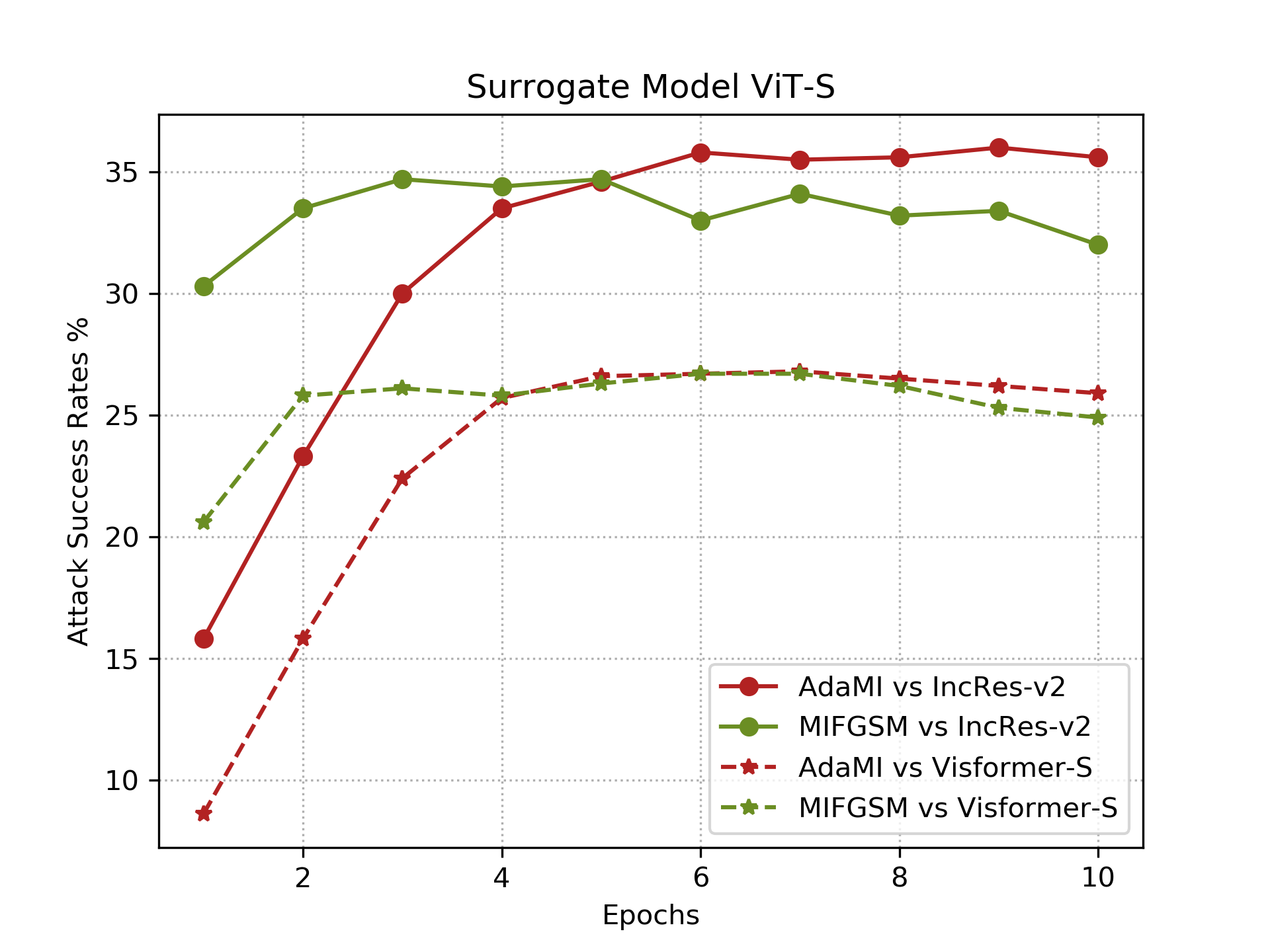}
\end{minipage}
\begin{minipage}{0.4\linewidth}
\centering
\includegraphics[width=\textwidth]{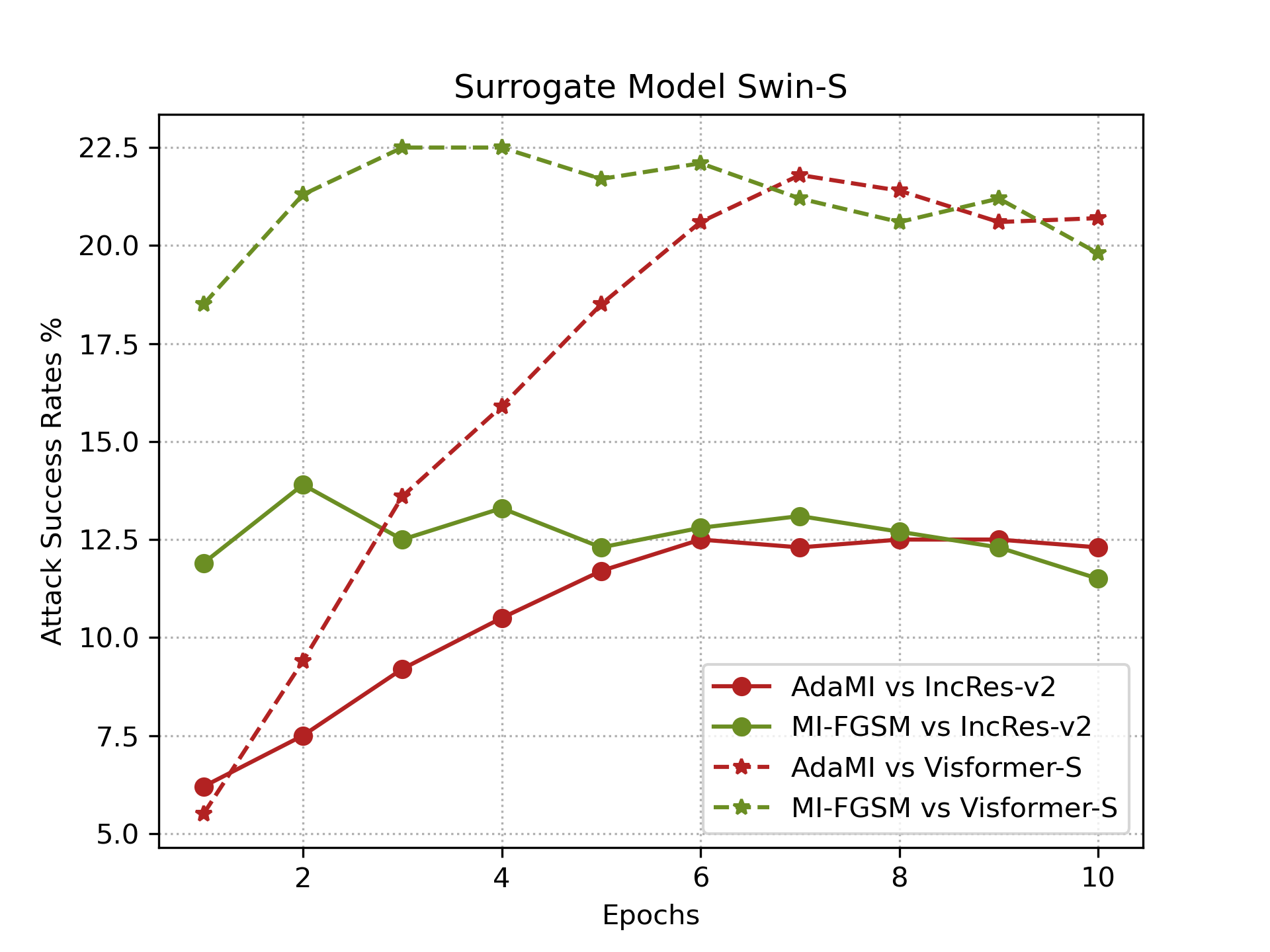}
\end{minipage}
\caption{\small{The success rates (\%) of the AEs under different numbers of epochs.}}
\label{fig:stability}
\end{figure*}

\subsection{Stability, Imperceptibility and Convergence}
\label{Stability}

To illustrate the stability of AdaMI, we show the correlation between success rates and the number of epochs. As depicted in Fig.\ref{fig:stability}, the integration of momentum into the iterative direction in MI-FGSM does contribute to stability. Notably, when $T$ is small, MI-FGSM might exhibit superior performance to AdaMI due to its larger step-size. Nonetheless, by further scaling this iterative direction with a momentum-based adaptive matrix, AdaMI can improve the transferability of MI-FGSM across different neural network architectures while maintaining better stability.

\begin{figure*}[htbp]
\centering
\begin{minipage}{0.4\linewidth}
\centering
\includegraphics[width=\textwidth]{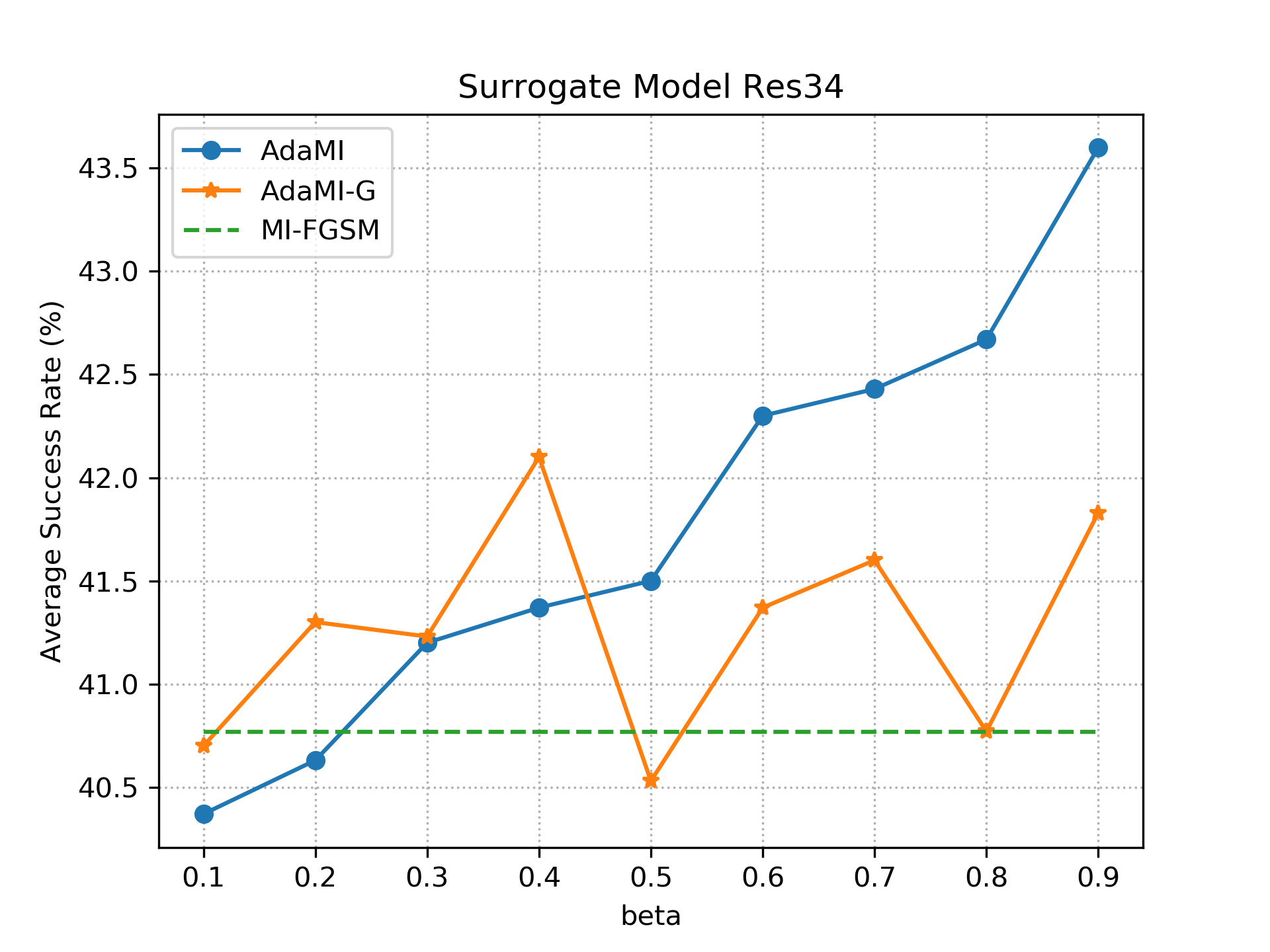}
\end{minipage}
\begin{minipage}{0.4\linewidth}
\centering
\includegraphics[width=\textwidth]{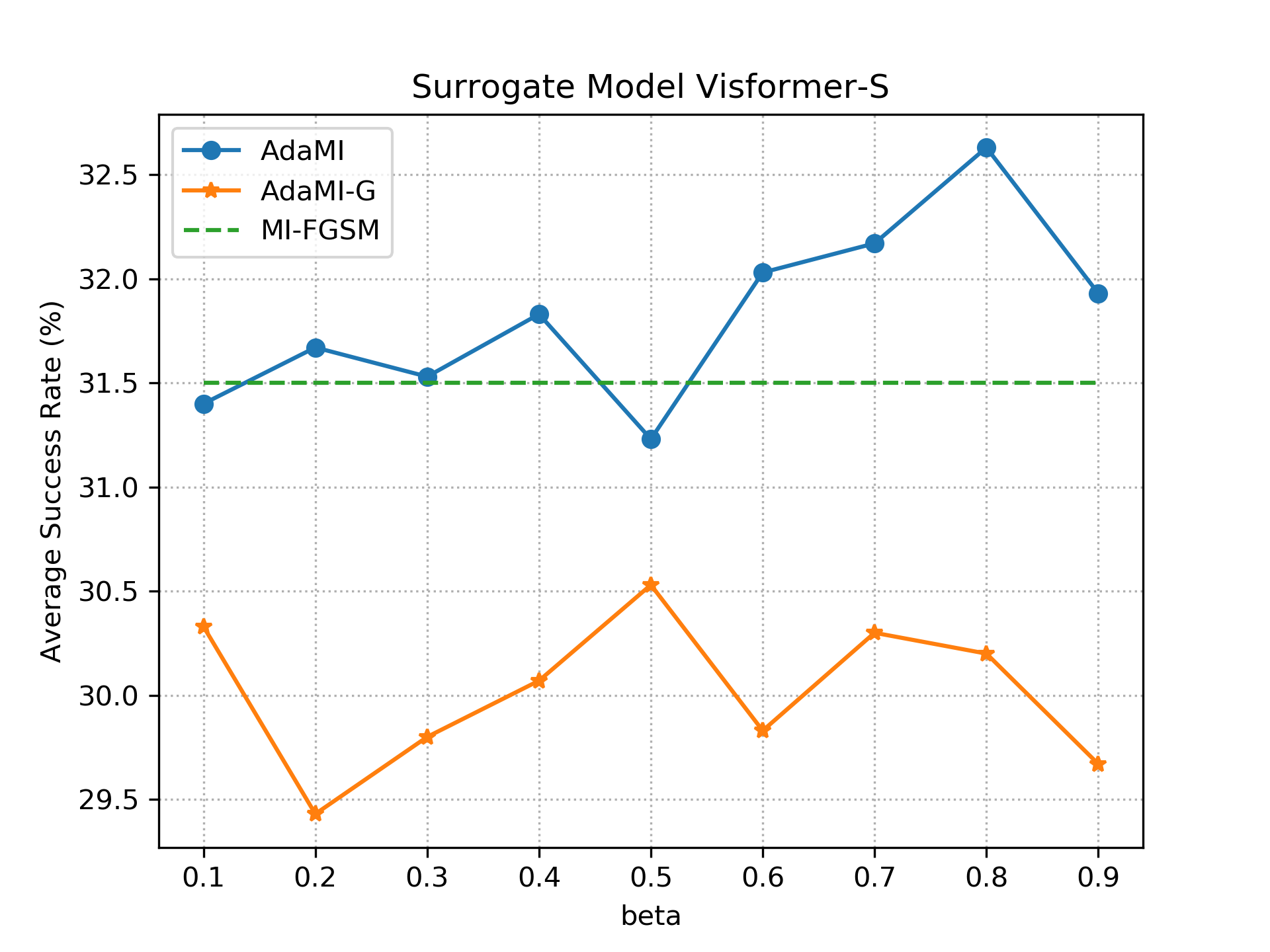}
\end{minipage}\\
\begin{minipage}{0.4\linewidth}
\centering
\includegraphics[width=\textwidth]{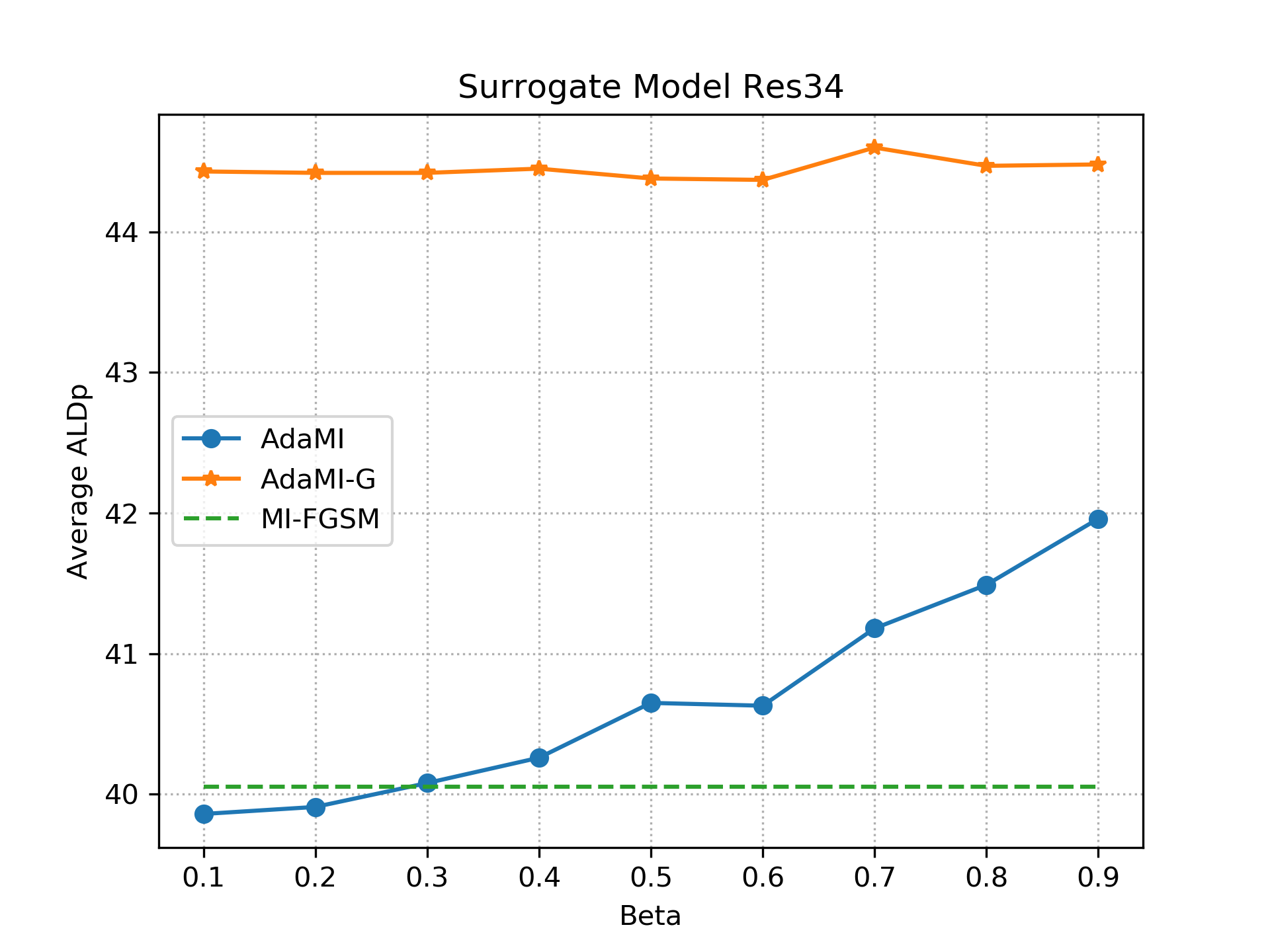}
\end{minipage}
\begin{minipage}{0.4\linewidth}
\centering
\includegraphics[width=\textwidth]{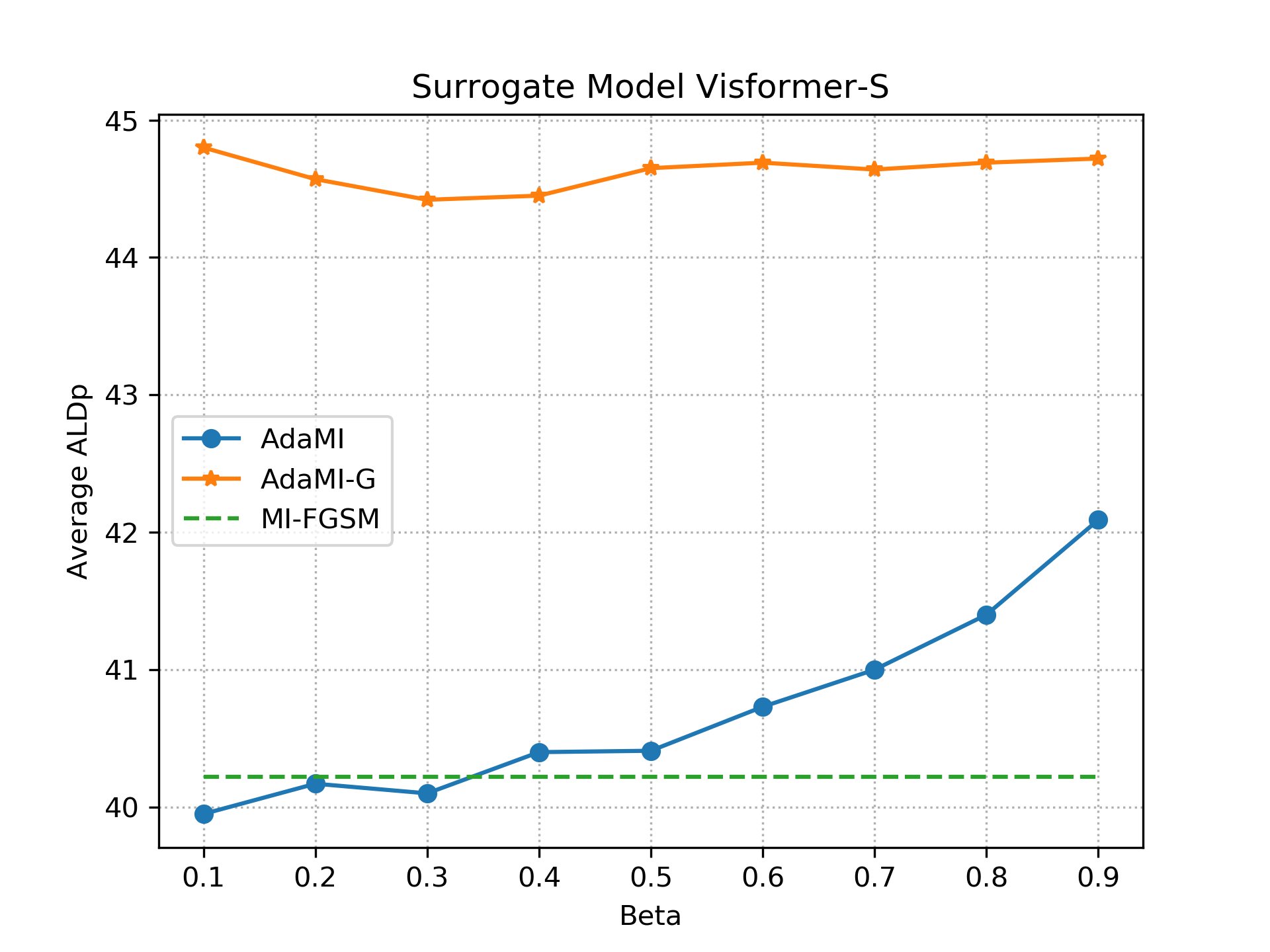}
\end{minipage}
\caption{Transferability and imperceptibility using different $\beta$. AEs are crafted for Res34 and Visformer-S, respectively.}
\label{fig:beta}
\end{figure*}

In recent years, optimization-based adversarial attacks have been developed rapidly. Although these methods have made significant progress in the transferability of adversarial attacks, but they have sacrificed the performance of imperceptibility. As can be seen, AutoAttack achieves the best results in terms of imperceptibility.
To highlight the benefits of the momentum-based adaptive matrix over its counterpart, we conducted an ablation study comparing AdaMI with MI-FGSM and AdaMI-G, in which different EMA parameters $\beta$ are used. As depicted in Fig.\ref{fig:beta}, 
AdaMI-G achieves a slightly lower success rate and imperceptibility among all three MI-based attack methods. By utilizing the momentum-based adaptive matrix, AdaMI can improve the transferability of MI-FGSM across different neural network architectures while maintaining a nearly equivalent level of imperceptibility. The performance of AdaMI validates the correctness of theoretical analysis and effectiveness of the algorithm. Consequently, the derived Ada-attacks with momentum-based adaptive matrix boost the transferability over the corresponding state-of-the-art methods.

\begin{figure*}[htbp]
\centering
\begin{minipage}{0.4\linewidth}
\centering
\includegraphics[width=\textwidth]{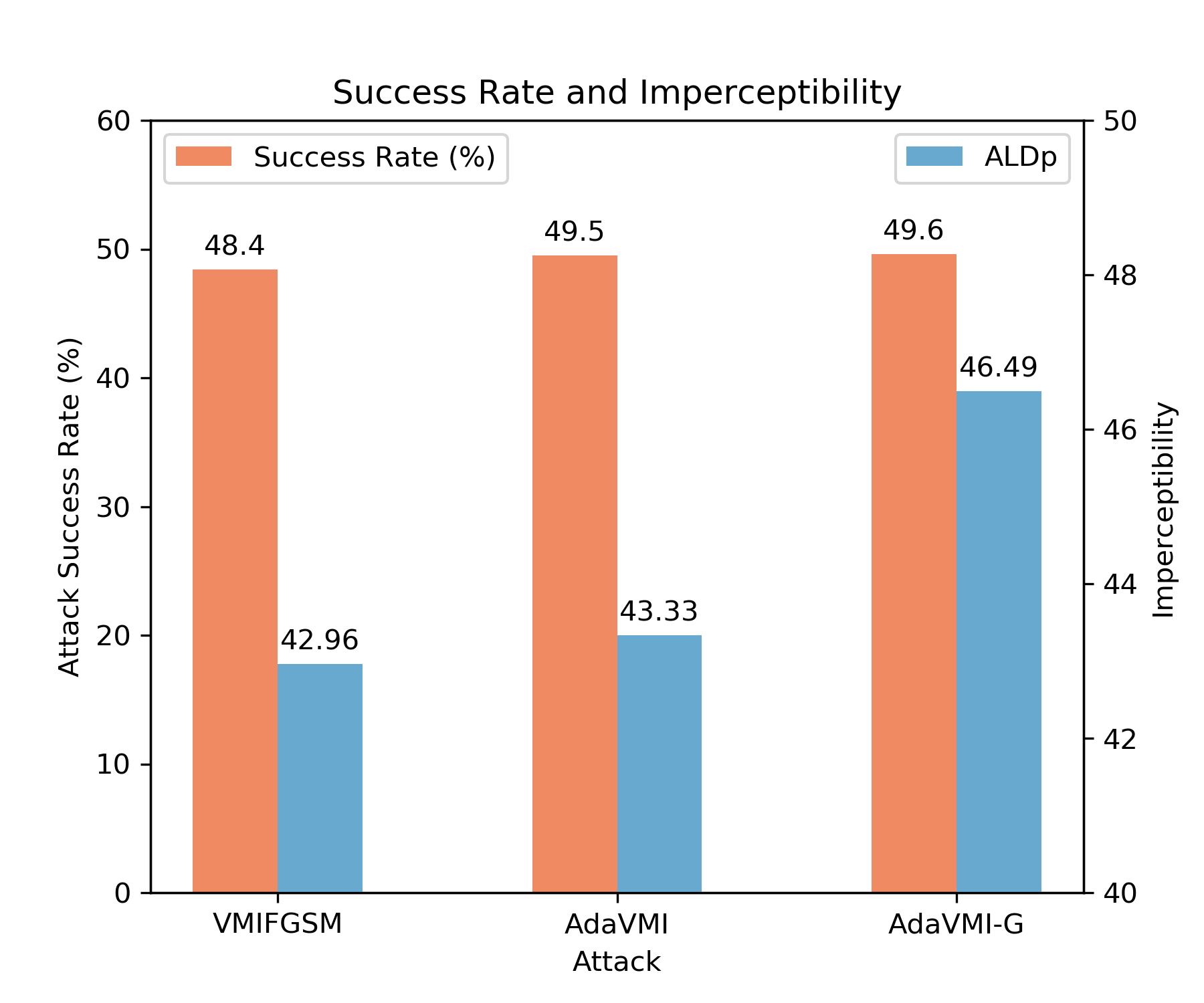}
\end{minipage}
\begin{minipage}{0.4\linewidth}
\centering
\includegraphics[width=\textwidth]{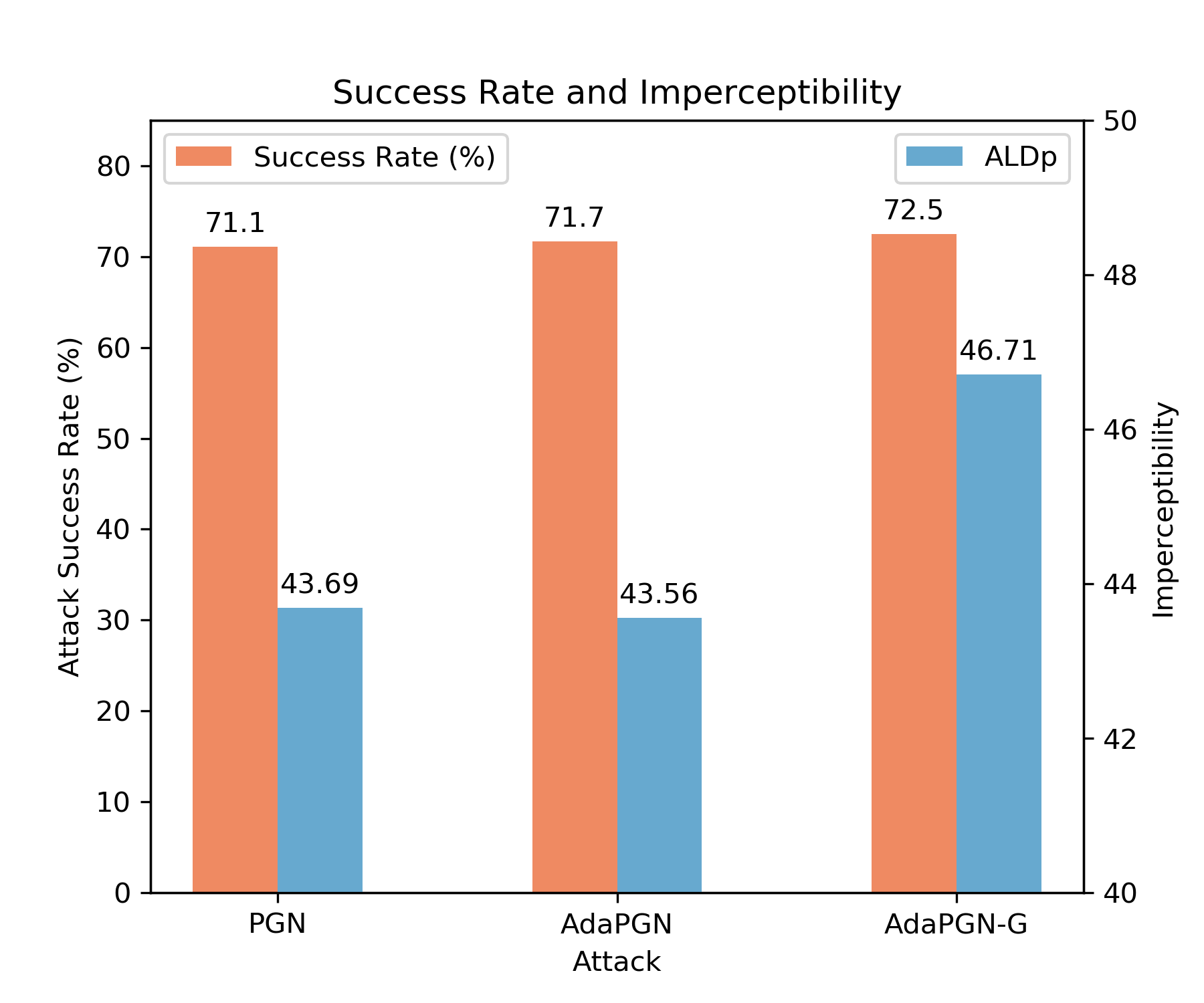}
\end{minipage}
\caption{Transferability and imperceptibility of gradient-based and momentum-based adaptive attacks. AEs are crafted for Res34.}
\label{fig:bar}
\end{figure*}

The comparison experiments between AdaMI and AdaMI-G have been conducted in Tab.\ref{tab:mom} and Fig.\ref{fig:beta}. Naturally, the gradient-based adaptive matrix can be readily combined with other momentum-based attacks (Tab.\ref{tab:momentum_all}). In practice, $v_{t+1,i}$ in AdaMI is usually larger than that in AdaMI-G, which can explain why the perturbations with smaller ALD$_p$ can be generated by AdaMI. Similar to AdaMI-G, we can easily derive AdaVMI-G and AdaPGN-G. To further illustrate the advantage of the momentum-based adaptive matrix over the gradient-based one, we compare respectively VMI and PGN with their adaptive variants, in which two different adaptive matrices are used. The attack success rates and ALD$_p$ are reported in Fig.\ref{fig:bar}.

From Fig.\ref{fig:bar}, it can be observed that AdaVMI-G and AdaPGN-G obtains slightly better transferability than AdaVMI and AdaPGN respectively. Unfortunately, in terms of the indicators of imperceptibility, AdaVMI-G and AdaPGN-G perform the worst among all the other attacks. Such a fact means that integrating the gradient-based adaptive matrix into other optimization-based attacks may slightly boost their transferability but at the cost of reduced imperceptibility. Nonetheless, generating momentum-based adaptive perturbation is capable of improving the transferability of vanilla momentum-based attacks while maintaining almost the same level of imperceptibility.

\begin{figure*}[htbp]
\centering
\begin{minipage}{0.42\linewidth}
\centering
\includegraphics[width=\textwidth]{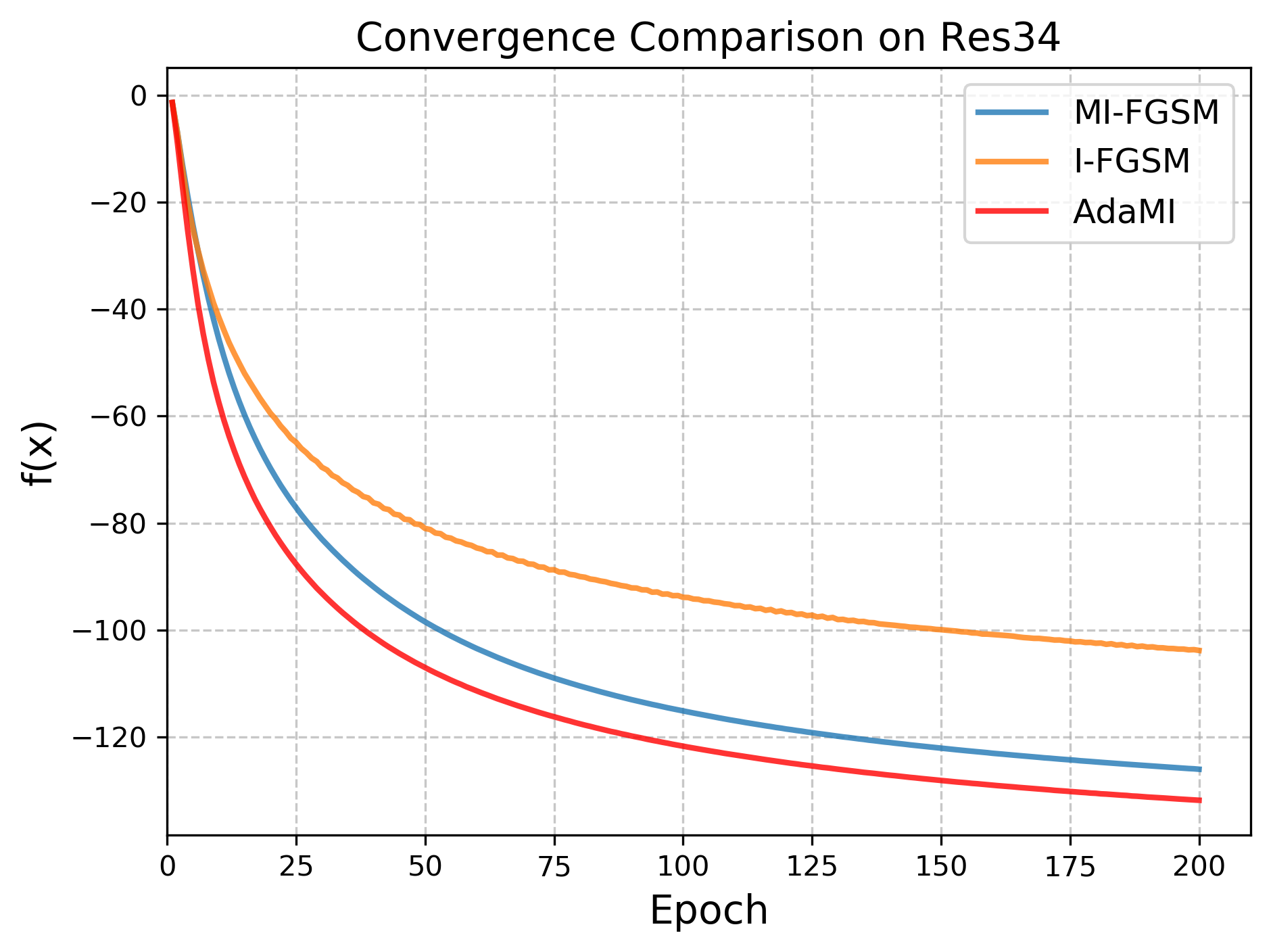}
\end{minipage}
\begin{minipage}{0.42\linewidth}
\centering
\includegraphics[width=\textwidth]{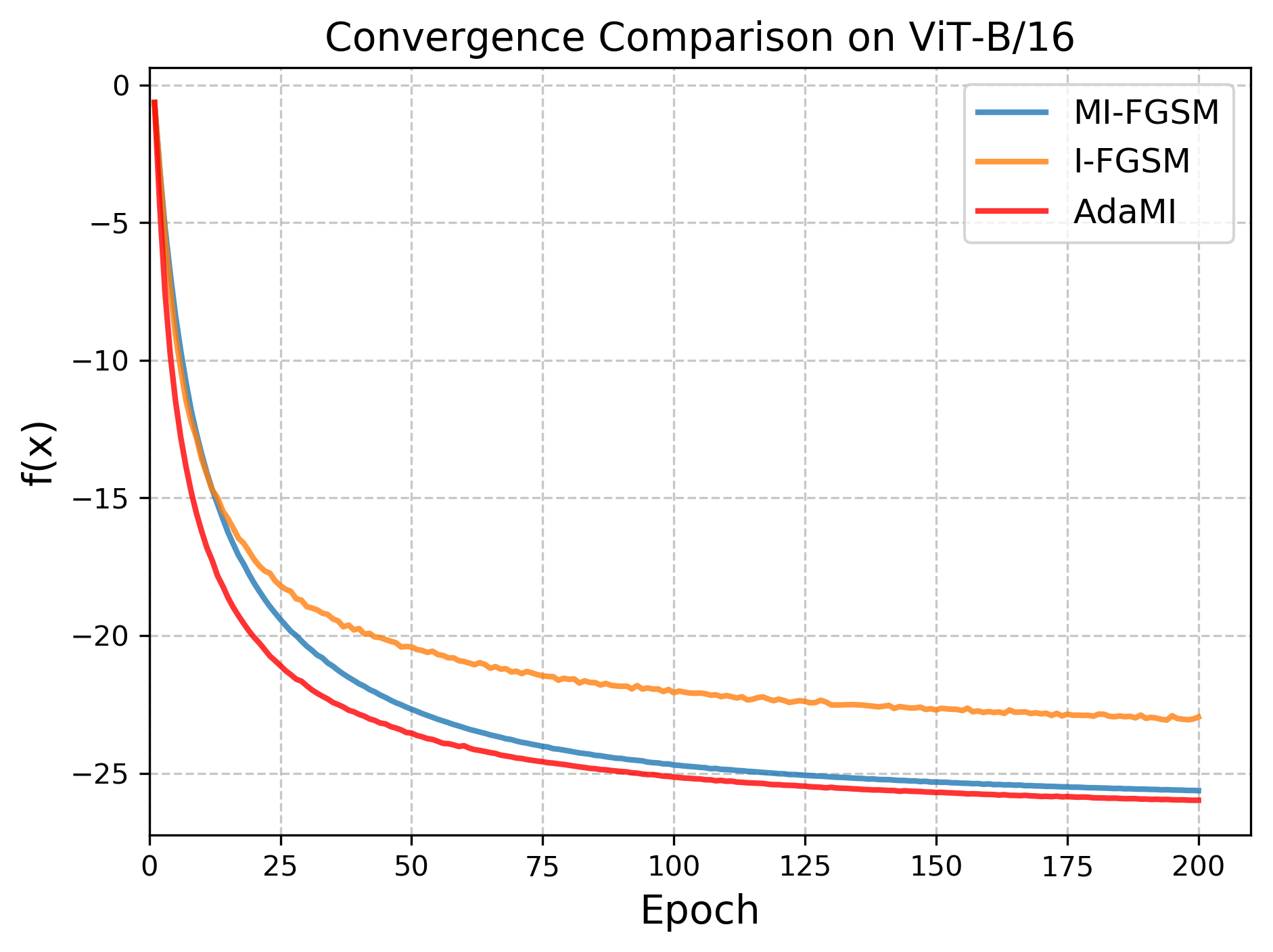}
\end{minipage}
\caption{Values of loss function of the generated AEs, where $f(\bm{x}) = -J(\bm{x},y)$.}
\label{fig:convergence}
\end{figure*}

To make a through comparison between AdaMI and MI-FGSM, we also investigate the convergence behavior of loss function $J(\bm{x}_{t}^{adv},y)$ with respect to the number of iterations. The relationship between the value of loss function $J(\bm{x}_{t}^{adv},y)$ and the number of iterations is shown in Fig.\ref{fig:convergence}. As can be seen in Fig.\ref{fig:convergence}, AdaMI converges consistently fast than MI-FGSM. In viewpoint of pure optimization algorithms, AdaMI has better convergence showing that it is more suitable for solving the adversarial attack optimization problems than MI-FGSM.

The computational cost comparisons are given in Tab.\ref{tab:efficiency}. It can be observed that our adaptive methods have almost the same level of run-time and max-GPU memory as their baseline counterparts, which indicates that incorporating adaptive momentum-based matrix incurs almost no additional computational cost. This is crucial for their adoption in large-scale or real-time applications.

\begin{table}[htbp]\small
\captionsetup{font={normalsize}}
\setlength{\tabcolsep}{1.2mm}
\caption{\normalsize{Computational cost comparisons between different adversarial attack methods on Res34.}}
\label{tab:efficiency}
\centering
\begin{tabular}{c|cccc}%
\toprule 
Attack &Epoch& Batchsize& Time (s)& \makecell{Max GPU\\ memory (Mb)}\\	
\midrule 
PGD & 10 & 50 &21.3&3148.2\\
I-FGSM & 10 & 50 &20.7&3151.2\\
AdaGrad& 10 & 50 & 23.5& 3237.3\\ 
MI& 10 & 50 & 20.9& 3151.5\\
AdaMI (Ours)& 10 & 50 & 23.8& 3266.5\\
NCS& 10 & 50 & 288.4& 5428.5\\ 
AdaNCS (Ours)& 10 & 50 & 278.5& 5543.5\\
\hline
\end{tabular}
\end{table}

\subsection{Additional Experiments}

Our method could be used to attack ensemble models or be integrated with various input transformations. In this section, we also conduct several additional experiments, including the use of adversarially trained models, ensemble models and input transformations.

\subsubsection{Combining with Input Transformation}

Like other optimization-based attacks, our adaptive strategy can also be combined with the input transformation-based methods to improve the transferability of the generated AEs. To further demonstrate the effectiveness of the proposed AdaMI and AdaNCS, we integrate the adaptive strategy into several typical input transformations i.e., TI \cite{Dong2019EvadingDT}, DI \cite{Xie2019ImprovingTO} , SI \cite{Lin2020NesterovAG} and Admix \cite{Wang2021admix}. For simplicity, we only consider that the AEs are crafted for Res34. The success rates are reported in Tab.\ref{tab:input transformation}.

\begin{table*}[htbp]\small
\captionsetup{font={normalsize}}
\caption{\normalsize{The success rates (\%) of adversarial attacks against baseline models. * indicates the results on the white-box model.}}
\label{tab:input transformation}
\centering
\begin{tabular}{c|cccccccc}
\toprule 
Attack & Res34& Inc-v3& VGG16& Mob-v2& ViT-S& ConViT& Visformer-S& Swin-T\\	
\midrule 
TI & 99.9$^*$&23.4&32.1&29.2&11.9&8.4&11.0&16.8\\
TI+MI& 99.8$^*$&34.1&46.9&40.3& 18.5&11.1&16.8&23.5\\
TI+AdaMI& 100.0$^*$& 37.1& 52.3& 43.6& 20.5& 12.5& 18.2& 25.5\\
TI+NCS & 99.8$^*$ & 47.5 & 59.9 & 52.5 & 27.3 & 16.7 & 23.8 & 33.1 \\
TI+AdaNCS & 99.9$^*$ & \bf{50.7} & \bf{62.8} & \bf{55.5} & \bf{28.0} & \bf{17.5} & \bf{25.8} & \bf{35.3} \\
\hline
DI& 99.9$^*$& 35.6& 40.5& 39.1& 14.5& 10.0& 15.9& 21.4\\
DI+MI &100.0$^*$&54.9&64.4&59.3&26.9&20.7&31.1&35.9\\
DI+AdaMI& 100.0$^*$& 58.5& 65.9& 60.8&27.0& 21.1& 33.6& 38.0 \\
DI+NCS& 99.9$^*$ & 61.9 & 71.5 & 67.4 & 38.8 & 32.7 & 42.0 & 48.2 \\
DI+AdaNCS&99.9$^*$ & \bf{63.5} & \bf{74.5} & \bf{69.4} & \bf{40.8} & \bf{33.7} & \bf{43.8} & \bf{51.2}  \\  \hline
SI& 99.9$^*$& 29.6& 37.0& 35.1& 12.1& 8.2& 12.2& 18.8\\
SI+MI& 100.0$^*$&52.6&60.3&53.9&21.3&15.8&24.3&33.4\\
SI+AdaMI& 100.0$^*$& 52.6& 62.0& 57.4& 22.5& 15.8& 25.4& 33.4\\ 
SI+NCS&100.0$^*$ & 62.8 & 75.5 & 71.1 & 35.4 & 29.1 & 42.8 & 49.7  \\
SI+AdaNCS& 100.0$^*$ & \bf{65.1} & \bf{79.1} & \bf{73.9} & \bf{36.8} & \bf{31.6} & \bf{44.2} & \bf{50.6} \\ 
\hline
Admix& 99.9$^*$& 37.9 & 49.5 & 46.2 & 15.5 & 10.8 & 16.7 &25.1 \\
Admix+MI& 100.0$^*$&63.0 & 71.7 & 65.4 & 27.6 & 20.9 & 33.3 & 40.6 \\
Admix+AdaMI& 100.0$^*$&62.8 & 72.8 & 66.4 & 28.0 & 21.5 & 34.7 & 41.7 \\ 
Admix+NCS&100.0$^*$ & 64.2 & 78.2 & 74.2 & 34.6 & 27.4& 41.2 & 50.8\\
Admix+AdaNCS& 100.0$^*$ &\bf{65.8} & \bf{78.6} & \bf{74.4} & \bf{34.8} & \bf{27.5} & \bf{41.5} & \bf{51.0}  \\ \hline
\end{tabular}
\end{table*}

\subsubsection{Attacking an Ensemble of Models}
\label{sec:4.3}

It has been shown that attacking multiple models at the same time can improve the transferability of generated AEs \cite{Liu2017DelvingIT}. To further compare with I-FGSM and MI-FGSM, we apply AdaGrad and AdaMI to attack an ensemble of models. As pointed out in \cite{Dong2018BoostingAA}, the ensemble in logits outperforms the ensemble in predictions and the ensemble in loss consistently among all the attack methods and different models in the ensemble for both the white-box and black-box attacks. Therefore, we only focus on attacking an ensemble of normally trained models in logits (including Inc-v3, Inc-v4 and IncRes-v2) with equal weights. SVRE \cite{Xiong2022svre} and AdaEA \cite{Chen2021AdaEA} are selected as baseline attacks in attacking ensemble models. We report the success rates of attack against different kinds of models in Tab.\ref{tab:ensemble}.

By comparing the experimental results in Tab.\ref{tab:gradient}, it is easy to find that the adaptive optimization-based attacks under the multi-model setting can similarly improve the transferability. Fortunately, our AdaMI consistently outperforms MI-FGSM and NI-FGSM when attacking an ensemble of models. The proposed AdaNCS achieves the best result.

\begin{table*}[htbp]\small
\captionsetup{font={normalsize}}
\caption{\normalsize{The success rates (\%) of adversarial attacks against ensemble models. * indicates the white-box model being attacked.}}
\label{tab:ensemble}
\centering
\begin{tabular}{c|cccccc}%
\toprule 
Attack&Inc-v3*& Inc-v4* &IncRes-v2* & Res-101& Inc-v3$_{adv}$& IncRes-v2$_{ens}$\\	
\midrule 
I-FGSM& 91.9$^*$& 84.4$^*$&80.1$^*$&12.3&26.2&13.6\\ 
AdaGrad & 94.2$^*$& 88.0$^*$&83.5$^*$&13.5&28.2& 14.5\\
SVRE & 94.4$^*$ & 92.2$^*$ & 91.1$^*$ &25.1 & 42.1 & 25.9 \\
AdaEA & 97.2$^*$ & 94.5$^*$ & 92.9$^*$ &29.8& 44.8 & 29.3 \\
NI& 97.3$^*$& 95.4$^*$&91.8$^*$&29.5&45.4&31.0\\
MI & 97.5$^*$& 95.7$^*$&93.1$^*$& 32.2& 48.0& 31.7\\
AdaMI (Ours) & 98.2$^*$& 96.6$^*$&96.2$^*$& 32.5& 48.0& 34.5\\
IE& 98.0$^*$& 97.8$^*$& 95.7$^*$& 35.8& 52.1&37.8\\
VMI& 97.6$^*$ & 95.9$^*$ & 92.2$^*$&41.3 &56.4 & 43.7\\
EMI& 98.7$^*$ & \bf{98.5$^*$} & \bf{98.2$^*$}&47.5 &64.1 & 49.7\\
MIG& 94.4$^*$ & 95.8$^*$ & 94.1$^*$ &42.7&  67.7& 54.7\\
RAP& 99.5$^*$ & 97.8$^*$ & 96.1$^*$ &50.1 & 67.0 & 49.6\\
GRA& 98.3$^*$& 96.9$^*$&95.0$^*$&54.5 & 70.3 & 60.0\\
PGN& \bf{99.1$^*$} & 97.8$^*$ & 95.2$^*$& 58.2&75.2 & 65.2\\ 
NCS& 97.6$^*$ & 95.5$^*$ & 93.0$^*$& 59.9 & 73.0 & 64.0\\
AdaNCS (Ours)& 98.6$^*$ & 97.4$^*$ & 95.5$^*$ &\bf{63.2}& \bf{77.7} & \bf{67.5} \\\hline
\end{tabular}
\end{table*}

\subsubsection{Attacking in Defense Scenarios}

To further verify the robustness of our AdaMI and AdaNCS, we evaluate the performance of the crafted AEs in defense scenarios. Following \cite{Yuan2022Natural} and \cite{Long2022Frequency}, we consider both defenses (AT, JPEG, HGD, RS, NRP) and adversarially trained models (Inception-v3$_{adv}$, Inception-ResNet-v2$_{ens}$, EfficientNet-B0, EfficientNet-B1). For simplicity, the AEs are crafted for Res34. The success rates are reported in Tab.\ref{tab:defense}.

\begin{table*}[htbp]\small
\captionsetup{font={normalsize}}
\caption{\normalsize{The success rates (\%) of adversarial attacks against defensed models or adversarially trained models. AEs are crafted for Res34.}}
\label{tab:defense}
\centering
\begin{tabular}{c|cccc|ccccc}%
\toprule 
Attack  &  Inc-v3$_{adv}$& IncRes-v2$_{ens}$& Efficient-B0 & Efficient-B1 &  AT& JPEG& HGD&RS &NRP\\	
\midrule 
I-FGSM & 16.7 & 8.7 & 10.3 & 8.9&32.1 & 30.9&5.2 &19.4 &25.3\\
AdaGrad & 16.4 & 9.5 & 11.1 & 9.2 &32.3 & 33.5&5.7 & 19.2&30.2\\
MI& 26.0&15.1&22.9&19.0& 32.9& 61.5& 16.9& 20.3&35.7\\
NI& 26.4&14.7&23.5&19.7&  32.9& 61.4& 16.6& 20.2&39.0\\
AdaMI (Ours)&26.7&15.3&23.7&21.6& 33.0& 62.8& 18.5& 20.3&39.3\\
VMI& 35.9 & 21.1 & 34.8 & 30.1& 33.2& 75.0&28.7 & 21.1&45.1\\
PI& 29.6 & 19.4 & 23.3 & 21.0 &34.3 & 56.0&17.3 & 23.2&43.0\\
GRA& 46.7 & 31.2 & 46.0 & 41.3 &35.5 & 81.0&39.9 & 24.3&59.8\\
NCS&51.1 & 35.4 & 52.2 & 47.0& 35.8& 84.8& 48.1& 24.6&61.4\\
AdaNCS (Ours)&\bf{53.0} & \bf{36.7} & \bf{54.3} & \bf{50.9}& \bf{35.8}& \bf{88.2}& \bf{48.1}& \bf{24.8}&\bf{62.2}\\\hline
\end{tabular}
\end{table*}

From Tab.\ref{tab:defense}, it can be observed that AdaMI outperforms MI-FGSM and NI-FGSM on almost all the defenses and adversarially trained models. Our AdaNCS also achieves the best results, which validate the effectiveness of the proposed Ada-attacks in attacking defense scenarios.

\subsubsection{Practical Applications}

\begin{figure}[htbp]
\centering
\includegraphics[width=0.45\textwidth]{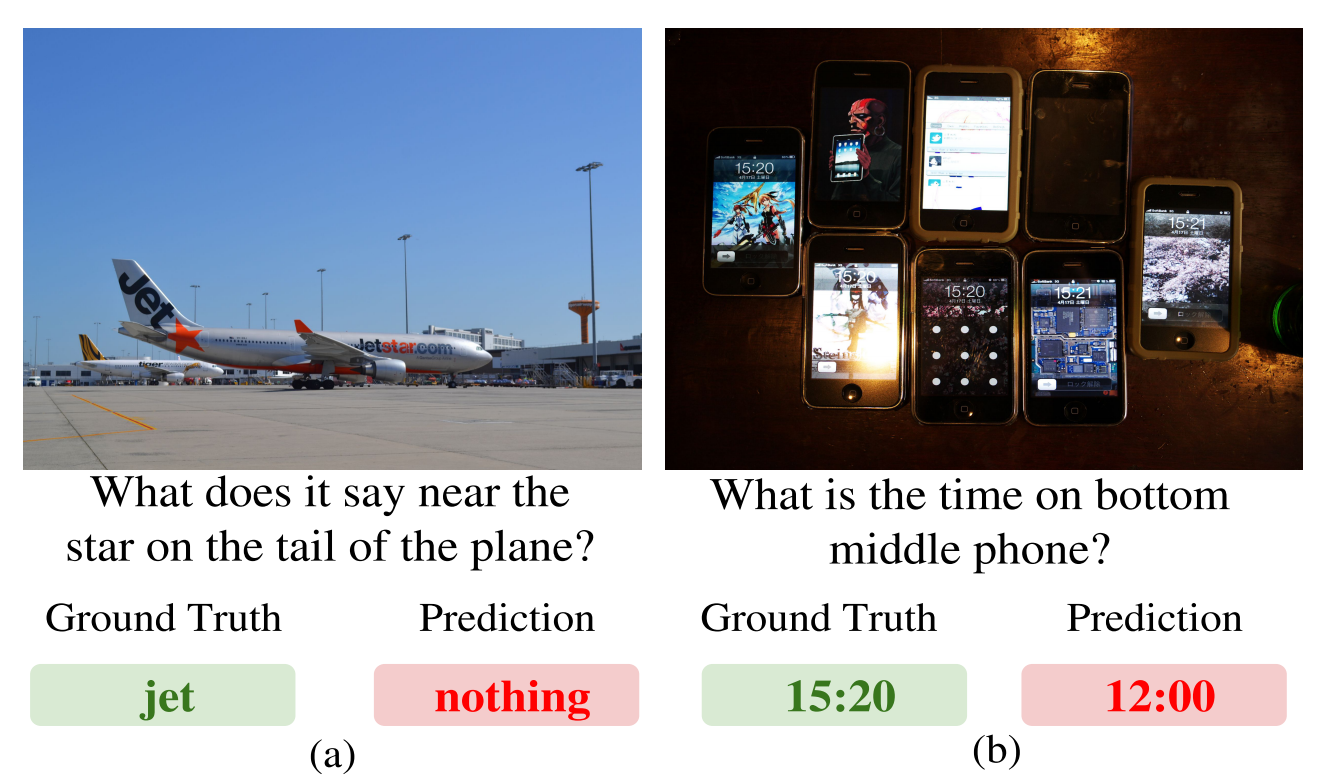}
\caption{Examples from TextVQA dataset \cite{vqa2020}.}
\label{fig:vqa}
\end{figure}

Visual Question Answering (VQA) serves as a practical application of Large Multimodal Models (LMMs), in which the model is tasked with generating an open-ended answer from a given image and an associated question. We evaluate the adversarial performance of our AdaMI using the LLaVA-1.5 \cite{llava} and PrismVLM \cite{Prism} models on the TextVQA dataset\footnote{\url{https://textvqa.org}}. Examples from TextVQA dataset can be found in Fig.\ref{fig:vqa}. The adversarial attack success rates are reported for two input types: Pure (raw textual questions) and OCR (augmented prompts that incorporate both raw questions and OCR-extracted text from the images). Unlike vision-only tasks, PGD is considered one of the state-of-the-art methods for attacking LMMs \cite{zhang2025vlms}. As shown in Tab.\ref{tab:vqa}, our AdaMI consistently achieves the lowest Post$_{N}$ values across all experimental settings, with particularly notable performance against the PrismVLM model.

\begin{table}[htbp]\small
\captionsetup{font={normalsize}}
\caption{\normalsize{The white-box adversarial attacks of Pure and OCR VQA tasks in LMMs. The notations Pre, Post$_{N}$ are refer to accuracy (\%) for pre-attack, post-attack under normal setting, respectively.}}
\label{tab:vqa}
\centering
\begin{tabular}{c|c|ccc}%
\toprule 
Model&Type& Attack& Pre& Post$_{N}$\\	
\midrule 
\multirow{10}{*}{LLaVA-1.5}&\multirow{5}{*}{Pure}& FGSM& 47.3& 37.2\\
& & PGD& 47.3& 22.1\\
&&MI&47.3&20.9\\
&& NCS& 47.3&22.1\\
&&AdaMI (Ours)&47.3&\bf{18.5}\\ \cline{2-5}
&\multirow{5}{*}{OCR}& FGSM& 58.5& 53.7\\
&& PGD& 58.5& 37.5\\
 & & MI& 58.5&37.2\\
&&NCS&58.5&36.3\\
&&AdaMI (Ours)&58.5&\bf{33.4}\\ \hline
\multirow{10}{*}{PrismVLM
}&\multirow{5}{*}{Pure}& FGSM& 56.9& 41.3\\
&& PGD& 56.9& 26.1\\
&&MI&56.9&26.0\\
&& NCS& 56.9&25.8\\
&&AdaMI (Ours)&56.9&\bf{21.5}\\ \cline{2-5}
&\multirow{5}{*}{OCR}& FGSM& 61.9& 49.7\\
&& PGD& 61.9& 31.4\\
&&MI&61.9&31.6\\
&& NCS& 61.9&31.1\\
&&AdaMI (Ours)&61.9&\bf{24.5}\\ \hline
\end{tabular}
\end{table}

\begin{figure*}[htbp]
\centering
\includegraphics[width=0.8\textwidth]{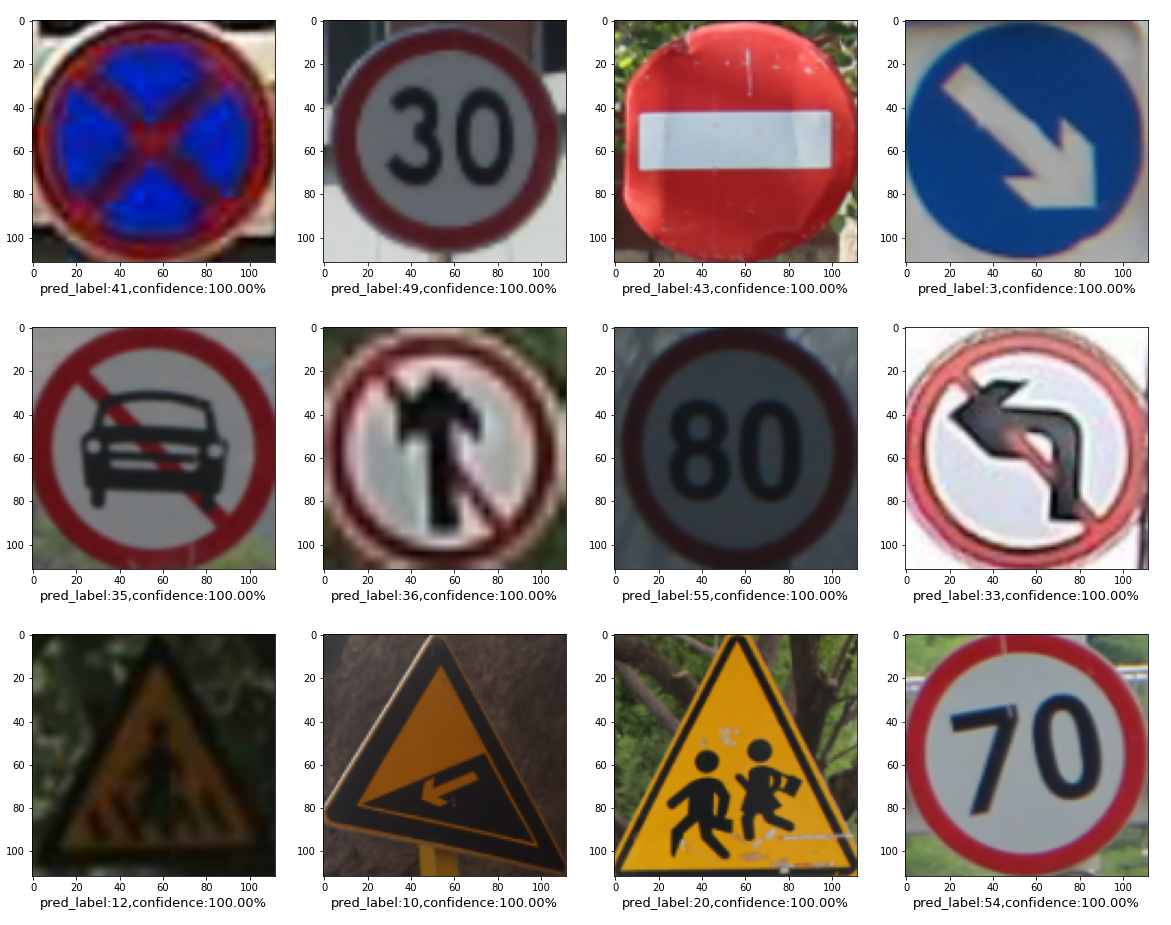}
\caption{Visualization of traffic signs with random transformations in CTSRD dataset. Classification model is Res34.}
\label{fig:traffic_clean}
\end{figure*}

\begin{figure*}[htbp]
\centering
\includegraphics[width=0.8\textwidth]{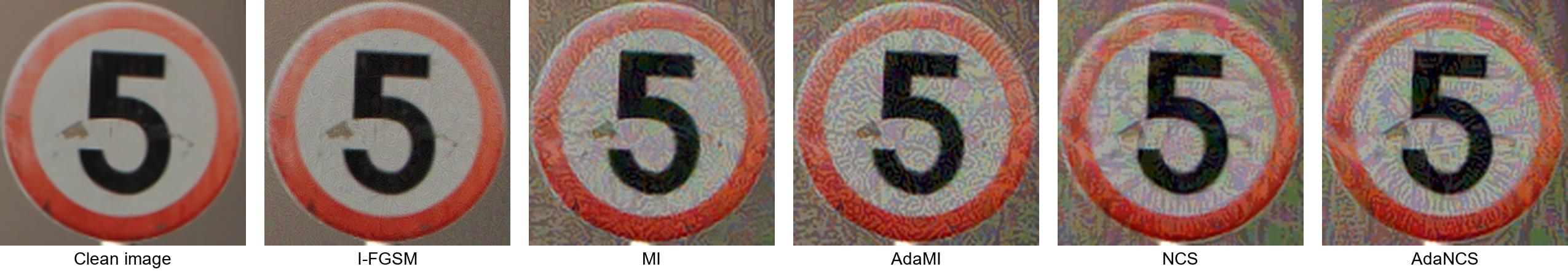}
\caption{Visualization of the clean image and its corresponding adversarial examples generated by different algorithms. We set the perturbation budget $\epsilon = 16/255$.}
\label{fig:traffic_adv}
\end{figure*}

Traffic sign recognition is a critical component of autonomous driving systems in real-world applications. Misclassification particularly when caused by a targeted attack on specific sign classes, can lead to severe and potentially catastrophic consequences. In this experiment, we evaluate the adversarial performance of the proposed algorithms using the Chinese Traffic Sign Recognition Database (CTSRD)\footnote{\url{https://nlpr.ia.ac.cn/pal/trafficdata/recognition.html}}. Visualization of traffic signs with random transformations can be seen in Fig.\ref{fig:traffic_clean}. As shown in Tab.\ref{tab:vqa}, our methods consistently achieve remarkably better results than their baseline counterparts. We also provide the original image of a 5 km/h speed limit traffic sign and its corresponding adversarial examples generated by different algorithms in Fig.\ref{fig:traffic_adv}.

\begin{table}[htbp]\small
\centering
\captionsetup{font={normalsize}}
\caption{Untargeted Attack Success Rate (\%) of traffic sign recognitions. AEs are crafted for Res34.}
\label{tab:taffic}
\begin{tabular}{c|c|ccc}
\toprule
   Model&Attack &  Res34&Res101&VGG16\\
\midrule
 &I-FGSM &  100.0$^*$&2.0& 4.6\\
 &MI &  100.0$^*$&8.2&23.9\\
 Res34&AdaMI (Ours) &  100.0*&9.5& 32.3\\
 &NCS& 100.0$^*$&12.8&32.7\\
 &AdaNCS (Ours)& 100.0$^*$&15.5&\bf{41.5}\\\hline
\end{tabular}
\end{table}

Note that the improvement of our proposed plug-and-play plugin is marginal in certain cases. This phenomenon can be understood from an optimization perspective. As formulated in optimization problem (1), adversarial attacks are essentially optimization problems aimed at misclassifying the model, typically using the cross-entropy loss. AdaMI is specifically designed to efficiently solve this type of optimization problem, making it particularly suitable for white-box attack settings. However, adversarial attacks are often evaluated in terms of their transferability, a property akin to generalization in machine learning, which is influenced by multiple factors beyond the optimization process, such as the choice of loss function and perturbation characteristics. For instance, recent studies \cite{Zhang2022ProvingCM} have shown that high transferability is negatively correlated with interaction among perturbation units, and a new objective function which jointly optimizes the classification loss and the interaction loss is introduced in \cite{Zhang2022ProvingCM} that can enhance transferability explicitly.

Although AdaMI does not consistently improve transferability in every case, especially when the attack already operates near the performance ceiling, it remains effective in the majority of evaluations, as demonstrated in our experiments. Moreover, AdaMI offers notable theoretical and practical advantages, such as improved stability and convergence speed, aligning with our core goal of introducing well-founded adaptive strategies into adversarial attack optimization, rather than solely pursuing empirical outperformance across all possible attacks and models.

\section{Conclusion}

In this paper, we address several theoretical concerns about optimization-based methods in adversarial attacks. We first analyze the connection between the basic gradient-based attack PGD and fundamental gradient method PGM. Motivated by such analysis, we present a new adaptive algorithm AdaMI based on the typical momentum-based attack MI-FGSM. The proposed AdaMI optimizes the adversarial perturbation with a novel momentum-based adaptive matrix, and it is proved to attain optimal convergence for convex problems which overcomes the issue of non-convergence of MI-FGSM. 

From the point of view of both theoretical analysis and empirical study, generating momentum-based adaptive perturbations can be serve as a general and effective technique to boost the adversarial transferability over the state-of-the-art methods across different neural network architectures while maintaining better stability and imperceptibility. As far as we known, the derived AdaNCS has achieved the best transferability among all the optimization-based attacks.


\bibliographystyle{IEEEtran}
\bibliography{egbib}

@book{bertsekas2003convex,
title={Convex analysis and optimization},
author={Dimitri P., Bertsekas and Angelia., Nedic and Asuman E., Ozdaglar},
year={2003},
publisher={Athena Scientific}
}

@inproceedings{Bernstein2018signSGDCO,
  title={signSGD: compressed optimisation for non-convex problems},
  author={Jeremy Bernstein and Yu-Xiang Wang and Kamyar Azizzadenesheli and Anima Anandkumar},
  booktitle={ICML},
  year={2018},
}

@inproceedings{Carlini2017TowardsET,
  title={Towards Evaluating the Robustness of Neural Networks},
  author={Nicholas Carlini and David A. Wagner},
  booktitle={IEEE SP},
  year={2017},
}

@article{Chawin2024,
  title={New Perspectives on Adversarially Robust Machine Learning Systems},
  author={Chawin Sitawarin},
  journal={Technical Report No. UCB/EECS-2024-10},
  year={2024}
 }

@inproceedings{Croce2020ReliableEO,
  title={Reliable evaluation of adversarial robustness with an ensemble of diverse parameter-free attacks},
  author={Francesco Croce and Matthias Hein},
  booktitle={ICML},
  year={2020},
 }

@inproceedings{Devlin2019BERTPO,
  title={BERT: Pre-training of Deep Bidirectional Transformers for Language Understanding},
  author={Jacob Devlin and Ming-Wei Chang and Kenton Lee and Kristina Toutanova},
  booktitle={NAACL},
  year={2019},
  }

@inproceedings{Dong2018BoostingAA,
  title={Boosting Adversarial Attacks with Momentum},
  author={Yinpeng Dong and Fangzhou Liao and Tianyu Pang and Hang Su and Jun Zhu and Xiaolin Hu and Jianguo Li},
  booktitle={CVPR},
  year={2018},
}

@inproceedings{Dong2019EvadingDT,
  title={Evading Defenses to Transferable Adversarial Examples by Translation-Invariant Attacks},
  author={Yinpeng Dong and Tianyu Pang and Hang Su and Jun Zhu},
  booktitle={CVPR},
  year={2019},
}

@inproceedings{Dong2020BenchmarkingAR,
  title={Benchmarking Adversarial Robustness on Image Classification},
  author={Yinpeng Dong and Qi-An Fu and Xiao Yang and Tianyu Pang and Hang Su and Zihao Xiao and Jun Zhu},
  booktitle={CVPR},
  year={2020},
}

@inproceedings{Goodfellow2015ExplainingAH,
  title={Explaining and Harnessing Adversarial Examples},
  author={Ian J. Goodfellow and Jonathon Shlens and Christian Szegedy},
  booktitle={ICLR},
  year={2015},
}

@inproceedings{Ge2023Boosting,
  title = {Boosting Adversarial Transferability by Achieving Flat Local Maxima},
  author = {Zhijin Ge and Xiaosen Wang and Hongying Liu and Fanhua Shang and Yuanyuan Liu},
  booktitle = {NeurIPS},
  year = {2023}
  }

@article{Gu2023survey,
title={A Survey on Transferability of Adversarial Examples across Deep Neural  Networks},
author= {Jindong Gu and Xiaojun Jia and  Pau de Jorge and  Wenqian Yu and Xinwei Liu and Avery Ma and Yuan Xun and Anjun Hu and  Ashkan Khakzar and Zhijiang Li and Xiaochun Cao and  Philip H. S. Torr},
journal= {ArXiv},
year= {2023},
volume={abs/2310.17626}
}

@inproceedings{Madry2018TowardsDL,
  title={Towards Deep Learning Models Resistant to Adversarial Attacks},
  author={Aleksander Madry and Aleksandar Makelov and Ludwig Schmidt and Dimitris Tsipras and Adrian Vladu},
  booktitle={ICLR},
  year={2018},
}

@inproceedings{Ma2023Transferable,
  title = {Transferable Adversarial Attack for Both Vision Transformers and Convolutional Networks via Momentum Integrated Gradients},
  author = {Wenshuo Ma and Yidong Li and  Xiaofeng Jia and  Wei Xu},
  booktitle    = {ICCV},
  year         = {2023},
  }

@inproceedings{Kingma2015AdamAM,
  title={Adam: A Method for Stochastic Optimization},
  author={Diederik P. Kingma and Jimmy Ba},
  booktitle={ICLR},
  year={2015},
  }

@inproceedings{krizhevsky2012imagenet,
  title={Imagenet classification with deep convolutional neural networks},
  author={Krizhevsky, Alex and Sutskever, Ilya and Hinton, Geoffrey E},
  booktitle={NIPS},
  year={2012}
}

@article{Kurakin2017AdversarialEI,
  title={Adversarial examples in the physical world},
  author={Alexey Kurakin and Ian J. Goodfellow and Samy Bengio},
  journal={ArXiv},
  year={2017},
  volume={abs/1607.02533}
}

@inproceedings{Li2023Improving,
  title={Improving Adversarial Transferability via Intermediate-level Perturbation Decay},
  author={Qizhang Li and Yiwen Guo and Wangmeng Zuo and Hao Chen},
  booktitle={NeurIPS},
  year={2023}
}

@inproceedings{Lin2020NesterovAG,
  title={Nesterov Accelerated Gradient and Scale Invariance for Adversarial Attacks},
  author={Jiadong Lin and Chuanbiao Song and Kun He and Liwei Wang and John E. Hopcroft},
   booktitle={ICLR},
  year={2020}
}

@inproceedings{Liu2017DelvingIT,
  title={Delving into Transferable Adversarial Examples and Black-box Attacks},
  author={Yanpei Liu and Xinyun Chen and Chang Liu and Dawn Xiaodong Song},
  booktitle={ICLR},
  year={2017}
  }

@inproceedings{Long2022Frequency,
  title={Frequency domain model augmentation for adversarial attack},
  author={Yuyang Long and Qilong Zhang and Boheng Zeng and Lianli Gao and Xianglong Liu and Jian Zhang and Jingkuan Song},
  booktitle={ECCV},
  year={2022}
  }

@inproceedings{mukkamala2017variants,
  title={Variants of rmsprop and adagrad with logarithmic regret bounds},
  author={Mukkamala, Mahesh Chandra and Hein, Matthias},
  booktitle={ICLR},
  year={2017}
}

@article{nesterov27method,
  title={A method of solving a convex programming problem with convergence rate $O(1/k^{2})$},
  author={Nesterov, Yu},
  journal={Soviet Mathematics Doklady},
  volume={27},
  number={2},
  pages={372--376},
  year={1983}
}

@article{Polyak1964SomeMO,
  title={Some methods of speeding up the convergence of iteration methods},
  author={Boris Polyak},
  journal={Ussr Computational Mathematics and Mathematical Physics},
  year={1964},
  volume={4},
  pages={1-17}
}

@inproceedings{reddi2019convergence,
  title={On the convergence of adam and beyond},
  author={Reddi, Sashank J and Kale, Satyen and Kumar, Sanjiv},
  booktitle={ICLR},
  year={2018}
}

@article{Russakovsky2015ImageNetLS,
  title={ImageNet Large Scale Visual Recognition Challenge},
  author={Olga Russakovsky and Jia Deng and Hao Su and Jonathan Krause and Sanjeev Satheesh and Sean Ma and Zhiheng Huang and Andrej Karpathy and Aditya Khosla and Michael S. Bernstein and Alexander C. Berg and Li Fei-Fei},
  journal={International Journal of Computer Vision},
  year={2015},
  volume={115},
  pages={211-252}
}

@inproceedings{LongAAAI,
  title={On the Convergence of an Adaptive Momentum Method for Adversarial Attacks},
  author = {Sheng Long and Wei Tao and Shuohao Li and Jun Lei and Jun Zhang},
  booktitle={AAAI},
  year= {2024}
}

@inproceedings{Tao2021TheRO,
  title={The Role of Momentum Parameters in the Optimal Convergence of Adaptive Polyak's Heavy-ball Methods},
  author={Wei Tao and Sheng Long and Gaowei Wu and Qing Tao},
  booktitle={ICLR},
  year={2021},
}

@article{tieleman2012lecture,
  title={Lecture 6.5-rmsprop, coursera: Neural networks for machine learning},
  author={Tieleman, Tijmen and Hinton, Geoffrey},
  journal={University of Toronto, Technical Report},
  year={2012}
}

@article{Xie2019ImprovingTO,
  title={Improving Transferability of Adversarial Examples With Input Diversity},
  author={Cihang Xie and Zhishuai Zhang and Jianyu Wang and Yuyin Zhou and Zhou Ren and Alan Loddon Yuille},
  journal={CVPR},
  year={2019},
  }

@inproceedings{Duchi2010AdaptiveSM,
  title={Adaptive Subgradient Methods for Online Learning and Stochastic Optimization},
  author={John C. Duchi and Elad Hazan and Yoram Singer},
  booktitle={J. Mach. Learn. Res.},
  year={2010}
}

@article{Ruder2016AnOO,
  title={An overview of gradient descent optimization algorithms},
  author={Sebastian Ruder},
  journal={ArXiv},
  year={2016},
  volume={abs/1609.04747}
}

@inproceedings{Karimireddy2019ErrorFF,
  title={Error Feedback Fixes SignSGD and other Gradient Compression Schemes},
  author={Sai Praneeth Karimireddy and Quentin Rebjock and Sebastian U. Stich and Martin Jaggi},
 booktitle={ICML},
  year={2019},
  }

@inproceedings{Peng2023Boosting,
  author={Anjie Peng and Zhi Lin and Hui Zeng and Wenxin Yu and Xiangui Kang},
  booktitle={ICASSP},
  title={Boosting Transferability of Adversarial Example via an Enhanced Euler's Method},
  year={2023}
  }

@article{Shi2020AdaptiveIA,
  title={Adaptive iterative attack towards explainable adversarial robustness},
  author={Yucheng Shi and Yahong Han and Quanxin Zhang and Xiaohui Kuang},
  journal={Pattern Recognit.},
  year={2020},
  volume={105},
  pages={107309}
}

@inproceedings{Szegedy2014IntriguingPO,
  title={Intriguing properties of neural networks},
  author={Christian Szegedy and Wojciech Zaremba and Ilya Sutskever and Joan Bruna and D. Erhan and Ian J. Goodfellow and Rob Fergus},
  booktitle={ICLR},
  year={2014},
}

@inproceedings{wang2019sadam,
  title={Sadam: A variant of adam for strongly convex functions},
  author={Wang, Guanghui and Lu, Shiyin and Tu, Weiwei and Zhang, Lijun},
  booktitle={ICLR},
  year={2020}
}

@inproceedings{Wang2021EnhancingTT,
  title={Enhancing the Transferability of Adversarial Attacks through Variance Tuning},
  author={Xiaosen Wang and Kun He},
  booktitle={CVPR},
  year={2021}
  }

@inproceedings{Wang2021BoostingTT,
  title={Boosting Adversarial Transferability through Enhanced Momentum},
  author={Xiaosen Wang and Jiadong Lin and Han Hu and Jingdong Wang and Kun He},
  booktitle={BMVC},
  year={2021}
  }

@article{Yang2022AdversarialEG,
  title={Adversarial example generation with AdaBelief Optimizer and Crop Invariance},
  author={Bo Yang and Hengwei Zhang and Yuchen Zhang and Kaiyong Xu and Jin-dong Wang},
  journal={ArXiv},
  year={2022},
  volume={abs/2102.03726}
}

@article{Yu2023ReliableEO,
  title={Reliable Evaluation of Adversarial Transferability},
  author={Wenqian Yu and Jindong Gu and Zhijiang Li and Philip H. S. Torr},
  journal={ArXiv},
  year={2023},
  volume={abs/2306.08565},
  }

@inproceedings{Yuan2022Natural,
  title={Natural color fool: Towards boosting black-box unrestricted attacks},
  author={Shengming Yuan and Qilong Zhang and Lianli Gao and Yaya Cheng and Jingkuan Song},
  booktitle={NeurIPS},
  year={2022},
}

@inproceedings{Zhang2022ProvingCM,
  title={Proving Common Mechanisms Shared by Twelve Methods of Boosting Adversarial Transferability},
  author={Quanshi Zhang and Xin Wang and Jie Ren and Xu Cheng and Shuyun Lin and Yisen Wang and Xiangming Zhu},
  booktitle={ICLR},
  year={2022},
  }

@inproceedings{Zhuang2020AdaBeliefOA,
  title={AdaBelief Optimizer: Adapting Stepsizes by the Belief in Observed Gradients},
  author={Juntang Zhuang and Tommy M. Tang and Yifan Ding and Sekhar C. Tatikonda and Nicha C. Dvornek and Xenophon Papademetris and James S. Duncan},
  booktitle={NeurIPS},
  year={2020},
}

@inproceedings{Zou2022MakingAE,
  title={Making Adversarial Examples More Transferable and Indistinguishable},
  author={Junhua Zou and Zhisong Pan and Junyang Qiu and Yexin Duan and Xin Liu and Yu Pan},
  booktitle={AAAI},
  year={2022}
}

@article{Yuan2024Adaptive,
  author       = {Zheng Yuan and  Jie Zhang and  Shiguang Shan},
  title        = {Adaptive Perturbation for Adversarial Attack},
  journal      = {ArXiv},
  volume       = {abs/2111.13841v3},
  year         = {2024},
  }

@article{Qiu2024Enhancing,
  author       = {Chunlin Qiu and Yiheng Duan and 
                  Lingchen Zhao and Qian Wang},
  title        = {Enhancing Adversarial Transferability 
                  Through neighborhood Conditional Sampling},
  journal      = {ArXiv},
  volume       = {abs/2405.16181},
  year         = {2024},
  }

@inproceedings{Liu2021swin,
  title={Swin transformer: Hierarchical vision transformer using shifted windows},
  author={Ze Liu and Yutong Lin and Yue Cao and Han Hu and Yixuan Wei and Zheng Zhang and Stephen Lin},
  booktitle={CVPR},
  year={2021},
  }

@inproceedings{Alexey2021Vit,
  title={An image is worth 16x16 words: Transformers for image recognition at scale},
  author={Alexey Dosovitskiy and Lucas Beyer and Alexander Kolesnikov and Dirk Weissenborn and Xiaohua Zhai},
  booktitle={ICLR},
  year={2021}
}

@inproceedings{Gao2020PI,
  title={Patch-wise attack for fooling deep neural network},
  author={Lianli Gao and Qilong Zhang and Jingkuan Song and Xianglong Liu and Heng Tao Shen},
  booktitle={ECCV},
  year={2020}
}

@inproceedings{Fang2024ANDA,
  title={Strong Transferable Adversarial Attacks via Ensembled Asymptotically Normal Distribution Learning},
  author={Zhengwei Fang and Rui Wang and Tao Huang and Liping Jing},
  booktitle={CVPR},
  year={2024}
}

@inproceedings{Zhu2024GRA,
  title={Boosting Adversarial Transferability via Gradient Relevance Attack},
  author={Hegui Zhu and Yuchen Ren and Xiaoyan Sui and  Lianping Yang and Wuming Jiang},
  booktitle={CVPR},
  year={2023}
}

@inproceedings{Zhu2024RAP,
  title={Boosting the Transferability of Adversarial Attacks with Reverse Adversarial Perturbation},
  author={Zeyu Qin and Yanbo Fan and Yi Liu and Li Shen and Yong Zhang and Jue Wang and Baoyuan Wu},
  booktitle={Neurips},
  year={2022}
}

@inproceedings{Martin2017FID,
  title={Gans trained by a two time-scale update rule converge to a local nash equilibrium},
  author={Martin Heusel and Hubert Ramsauer and Thomas Unterthiner and Bernhard Nessler and Sepp Hochreiter},
  booktitle={Neurips},
  year={2017}
}

@inproceedings{Wang2021admix,
  title={Admix: Enhancing the transferability of adversarial attacks},
  author={Xiaosen Wang and Xuanran He and Jingdong Wang and Kun He},
  booktitle={ICCV},
  year={2021}
}

@inproceedings{Chen2021AdaEA,
  title={An adaptive model ensemble adversarial attack for boosting adversarial transferability},
  author={Bin Chen and Jia-Li Yin, Shukai Chen and Bo-Hao Chen and Ximeng Liu},
  booktitle={ICCV},
  year={2023}
}

@inproceedings{Xiong2022svre,
  title={Stochastic Variance Reduced Ensemble Adversarial Attack for Boosting the Adversarial Transferability},
  author={Yifeng Xiong and Jiadong Lin and Min Zhang and John E. Hopcroft and Kun He},
  booktitle={CVPR},
  year={2022}
}

@article{yang2025A,
  title={Adversarial example soups: Improving transferability and stealthiness for free},
  author={Yang, Bo and Zhang, Hengwei and Wang, Jindong and Yang, Yulong and Lin, Chenhao and Shen, Chao and Zhao, Zhengyu},
  journal={IEEE Transactions on Information Forensics and Security},
  year={2025},
  publisher={IEEE}
}

@article{zhang2025explainable,
  title={Explainable and transferable adversarial attack for ML-based network intrusion detectors},
  author={Zhang, Hangsheng and Han, Dongqi and Zhuang, Shangyuan and Wang, Zhiliang and Sun, Jiyan and Liu, Yinlong and Liu, Jiqiang and Dong, Jinsong},
  journal={IEEE Transactions on Dependable and Secure Computing},
  year={2025},
  publisher={IEEE}
}

@inproceedings{zhang2018camou,
  title={CAMOU: Learning physical vehicle camouflages to adversarially attack detectors in the wild},
  author={Zhang, Yang and Foroosh, Hassan and David, Philip and Gong, Boqing},
  booktitle={International Conference on Learning Representations},
  year={2018}
}

@article{huang2023erosion,
  title={Erosion attack: Harnessing corruption to improve adversarial examples},
  author={Huang, Lifeng and Gao, Chengying and Liu, Ning},
  journal={IEEE Transactions on Image Processing},
  volume={32},
  pages={4828-4841},
  year={2023},
  publisher={IEEE}
}

@inproceedings{vqa2020,
  title={Towards vqa models that can read},
  author={Amanpreet Singh and Vivek Natarajan and Meet Shah and Yu Jiang and Xinlei Chen and Dhruv Batra and Devi Parikh and Marcus Rohrbach},
  booktitle={CVPR},
  year={2019},
  pages={8317-8326}
}

@inproceedings{Prism,
  title={Prismatic vlms: Investigating the design space of visually-conditioned language models},
  author={Siddharth Karamcheti and Suraj Nair and Ashwin Balakrishna and Percy Liang and Thomas Kollar and Dorsa Sadigh},
  booktitle={ICLR},
  year={2024}
}

@inproceedings{llava,
  title={Improved baselines with visual instruction tuning},
  author={Haotian Liu and Chunyuan Li and Yuheng Li and Yong Jae Lee},
  booktitle={CVPR},
  year={2024},
  pages={26296-26306}
}

@inproceedings{zhang2025vlms,
  title={On the robustness of large multimodal models against image adversarial attacks},
  author={Xuanming Cui and Alejandro Aparcedo and Young Kyun Jang and Ser-Nam Lim},
  booktitle={CVPR},
  year={2024},
  pages={24625-24634}
}

\clearpage
\begin{appendix}
\subsection{Convergence Analysis of AdaMI}
\label{sec:appendix}

\begin{lemma}
\label{lem:usefullemma1}
Let $1>\lambda>0$ and $\mu>0$. Let $\mu_t=\mu \lambda^{t-1}$.
Suppose $\{\bm{g}_{t+1}\}_{t=1}^{\infty}$ is generated by AdaMI. Then there exits a $D_2>0$ such that
$$\|\bm{g}_{t+1}\| \leq D_2,\ \forall t>0,$$
\end{lemma}
{\it Proof} \ Note $\| \nabla_{\boldsymbol{x}}J(\boldsymbol{x}^{adv})\| \leq \| \nabla_{\boldsymbol{x}}J(\boldsymbol{x}^{adv})\|_{1}$. Then
$$ \|\bm{g}_{t+1}\|\leq \mu_t \|\bm{g}_{t}\|+1  \leq  \mu_{t-1} \|\bm{g}_{t-1}\|+ \mu_{t}+1 \leq \sum_{i=1}^{t} \mu_i +1.$$
Lemma \ref{lem:usefullemma1} follows from the convergence of $\sum_{i=1}^{t} \mu_i$.

In order to make our proof easy to understand, we first consider a specific AdaMI without using the adaptive matrix, which can be formulated as

\begin{equation}\label{Simple AdaMI}
\left  \{
\begin{array}{l}
\bm{g}_{t+1}=\mu_t \ \bm{g}_t  +  \displaystyle\frac{\nabla_{\bm{x}} J(\bm{x}_{t}^{adv})}{\|\nabla_{\bm{x}} J(\bm{x}_{t}^{adv})\|_1}\\
\bm{x}_{t+1}^{adv}= P_{\mathbf{Q}} \left(\bm{x}_{t}^{adv} +\alpha_t \bm{g}_{t+1}\right)
\end{array},
\right.
\end{equation}
where $\alpha_t=\displaystyle\frac{\alpha}{\sqrt{t}}$ and  $\mu_t=\mu \lambda^{t-1}$ are as defined in Lemma \ref{lem:usefullemma1}.

\begin{lemma}
\label{lem:Simple AdaMI}
Let $1>\lambda>0$, $\mu>0$ and $\alpha>0$. Let $\mu_t=\mu \lambda^{t-1}$ and $\alpha_t= \displaystyle\frac{\alpha }{\sqrt{t}}$.
Let $\{ \boldsymbol{x}^{adv}_t\}_{t=1}^{\infty} $ be generated by AdaMI (\ref{Simple AdaMI}). Then we have
  $$J(\boldsymbol{x}^\ast)-J(\boldsymbol{\bar{x}}_T^{adv}) \leq O(\frac{1}{\sqrt T}).$$
\end{lemma}

{\it Proof} \ \ From the non-expansiveness of the projection operator, we know
\begin{equation*}
   \begin{aligned}
   & \|\bm{x}_{t+1}^{adv}-\boldsymbol{x}^{\ast}\|^2 \\
   &\leq \|\bm{x}_{t}^{adv}+\alpha_t \boldsymbol{g}_{t+1} -\boldsymbol{x}^{\ast}\|^2 \\
   & = \|\bm{x}_{t}^{adv}-\boldsymbol{x}^{\ast}\|^2 + \| \alpha_t \boldsymbol{g}_{t+1} \|^2 \\
   & \ \ \ \ +2\alpha_t \langle \boldsymbol{g}_{t+1} , \bm{x}_{t}^{adv}-\boldsymbol{x}^{\ast}\rangle \\
   & = \|\bm{x}_{t}^{adv}-\boldsymbol{x}^{\ast}\|^2 + \| \alpha_t \boldsymbol{g}_{t+1} \|^2  \\
   & \ \ \ \ + 2\alpha_t\langle \mu_t \boldsymbol{g}_{t}+ \displaystyle\frac{\nabla_{\bm{x}} J(\bm{x}_{t}^{adv})}{\|\nabla_{\bm{x}} J(\bm{x}_{t}^{adv})\|_1}, \bm{x}_{t}^{adv}-\boldsymbol{x}^{\ast}\rangle.
   \end{aligned}
 \end{equation*}
 Rearrange the inequality, we have
  \begin{equation*}
  \begin{aligned}
   &\frac{2\alpha_t}{\|\nabla_{\bm{x}} J(\bm{x}_{t}^{adv})\|_1} \langle\nabla_{\bm{x}} J(\bm{x}_{t}^{adv}), \bm{x}_{t}^{adv}-\boldsymbol{x}^{\ast}\rangle \\
   & \geq  \|\bm{x}_{t+1}^{adv}-\boldsymbol{x}^{\ast}\|^2 -\|\bm{x}_{t}^{adv}-\boldsymbol{x}^{\ast}\|^2  - \| \alpha_t \boldsymbol{g}_{t+1} \|^2  \\
   & \ \ \ \ - 2\alpha_t \mu_t \langle\boldsymbol{g}_{t} , \bm{x}_{t}^{adv}-\boldsymbol{x}^{\ast}\rangle,
   \end{aligned}
 \end{equation*}
 i.e.,
  \begin{equation*}
  \begin{aligned}
 & \frac{1}{\|\nabla_{\bm{x}} J(\bm{x}_{t}^{adv})\|_1} \langle\nabla_{\bm{x}} J(\bm{x}_{t}^{adv}), \bm{x}_{t}^{adv}-\boldsymbol{x}^{\ast}\rangle\\
 & \geq  \frac{\|\bm{x}_{t+1}^{adv}-\boldsymbol{x}^{\ast}\|^2 -\|\bm{x}_{t}^{adv}-\boldsymbol{x}^{\ast}\|^2}{2\alpha_t } - \frac{\alpha_t \|\boldsymbol{g}_{t+1} \|^2}{2} \\
 & \ \ \ \ -\mu_t \langle\boldsymbol{g}_{t} , \bm{x}_{t}^{adv}-\boldsymbol{x}^{\ast}\rangle \\
 & \geq  \frac{\|\bm{x}_{t+1}^{adv}-\boldsymbol{x}^{\ast}\|^2 -\|\bm{x}_{t}^{adv}-\boldsymbol{x}^{\ast}\|^2}{2\alpha_t } - \frac{\alpha_t \|\boldsymbol{g}_{t+1} \|^2}{2}  \\
 & \ \ \ \ - \frac{\mu_t \alpha_t \| \boldsymbol{g}_{t} \|^2}{2 } - \frac{ \mu_t \|\bm{x}_{t}^{adv}-\boldsymbol{x}^{\ast}\|^2}{2 \alpha_t}. \\
 \end{aligned}
 \end{equation*}
 Using the property of concave functions,
 \begin{equation*}
   \langle\nabla_{\bm{x}} J(\bm{x}_{t}^{adv}), \bm{x}_{t}^{adv}-\boldsymbol{x}^{\ast}\rangle
  \leq J(\boldsymbol{x}^{adv}_{t})-J(\boldsymbol{x}^{\ast}).
 \end{equation*}
 Then
\begin{equation*}
\begin{aligned}
& \frac{J(\boldsymbol{x}^{\ast})-J(\boldsymbol{x}^{adv}_{t})}{G} \\
& \leq \frac{\|\bm{x}_{t}^{adv}-\boldsymbol{x}^{\ast}\|^2 - \|\bm{x}_{t+1}^{adv}-\boldsymbol{x}^{\ast}\|^2 }{2\alpha_t } \\
& \ \ \ \ + \frac{\alpha_t \|\boldsymbol{g}_{t+1} \|^2}{2} + \frac{\mu_t \alpha_t\| \boldsymbol{g}_{t} \|^2}{2 } + \frac{ \mu_t \|\bm{x}_{t}^{adv}-\boldsymbol{x}^{\ast}\|^2}{2 \alpha_t}. \\
\end{aligned}
\end{equation*}
 Summing this inequality from $t = 1$ to $T$, we obtain
\begin{equation*}
\begin{aligned}
& \frac{1}{G}\sum_{t=1}^{T}\left[J(\boldsymbol{x}^{\ast})-J(\boldsymbol{x}^{adv}_{t})\right] \\
&  \leq\underbrace{\sum_{t=1}^{T}\frac{\|\bm{x}_{t}^{adv}-\boldsymbol{x}^{\ast}\|^2 - \|\bm{x}_{t+1}^{adv}-\boldsymbol{x}^{\ast}\|^2 }{2\alpha_t }}_{P_1} \\
& \ \ \ \ +\underbrace{\sum_{t=1}^{T}\frac{ \mu_t \|\bm{x}_{t}^{adv}-\boldsymbol{x}^{\ast}\|^2}{2 \alpha_t}}_{P_2}\\
&\ \ \ \ +\underbrace{\sum_{t=1}^{T}\frac{\alpha_t \|\boldsymbol{g}_{t+1} \|^2}{2}+\sum_{t=1}^{T}\frac{\mu_t \alpha_t\| \boldsymbol{g}_{t} \|^2}{2 }.}_{P_3}
\end{aligned}
\end{equation*}

To bound $P_1$, we have
\begin{equation}\label{c}
\begin{aligned}
     P_1=&\sum_{t=1}^{T}\frac{\|\bm{x}_{t}^{adv}-\boldsymbol{x}^{\ast}\|^2 - \|\bm{x}_{t+1}^{adv}-\boldsymbol{x}^{\ast}\|^2 }{2\alpha_t }\\
     =&\sum_{t=2}^{T}\left(\frac{1}{2\alpha_t}-\frac{1}{2\alpha_{t-1}}\right)\|\boldsymbol{x}_{t}^{adv}-\boldsymbol{x}^{\ast}\|^2 \\ 
     & + \frac{\|\boldsymbol{x}_{1}^{adv}-\boldsymbol{x}^{\ast}\|^2}{2\alpha_1}
     -\frac{\|\boldsymbol{x}_{T+1}^{adv}-\boldsymbol{x}^{\ast}\|^2}{2\alpha_T }\\
     \leq & \sum_{t=2}^{T}\left(\frac{1}{2\alpha_t}-\frac{1}{2\alpha_{t-1}}\right){D_1}^2 +\frac{D_{1}^2}{2\alpha_1} \\
     = & \frac{D_{1}^2}{2\alpha_T}  \leq \frac{D_{1}^2\sqrt{T}}{2\alpha }.
\end{aligned}
\end{equation}
To bound $P_2$, we have
\begin{equation}\label{d}
\begin{aligned}
P_2 & =\sum_{t=1}^{T}\frac{ \mu_t \|\bm{x}_{t}^{adv}-\boldsymbol{x}^{\ast}\|^2}{2 \alpha_t} \\
    & \leq\frac{{\mu D_{1}}^2}{2\alpha}\sum_{t=1}^{T} \lambda^{t-1} \sqrt{t} \\
    & \leq \frac{{\mu D_{1}}^2}{2\alpha}\sum_{t=1}^{T} \lambda^{t-1} t \\
    & \leq \frac{{\mu D_{1}}^2}{2\alpha (1-\lambda)^2}
\end{aligned}
\end{equation}
 To bound $P_3$, according to Lemma \ref{lem:usefullemma1}, we have
  \begin{equation}\label{e}
   \begin{aligned}
     P_3&=\sum_{t=1}^{T}\frac{\alpha_t \|\boldsymbol{g}_{t+1} \|^2}{2}+\sum_{t=1}^{T}\frac{\mu_t \alpha_t\| \boldsymbol{g}_{t} \|^2}{2 } \\
     & \leq \sum_{t=1}^{T} \frac{\alpha_t {D_{2}}^2}{2}+\sum_{t=1}^{T}\frac{\mu_t \alpha_t {D_{2}}^2}{2 }\\
     &\leq \frac{D_{2}^2}{2} \sum_{t=1}^{T} \frac{\alpha}{\sqrt t} +\frac{D_{2}^2}{2} \sum_{t=1}^{T}\frac{\mu_t \alpha}{\sqrt t} \\
     & \leq 2\alpha {D_{2}}^2 \sqrt{T}.
   \end{aligned}
\end{equation}
Combining \ref{c}, \ref{d} and \ref{e}, we have
\begin{equation*}
\begin{aligned}
    & \frac{1}{G} \sum_{t=1}^{T}\left( J(\boldsymbol{x}^{\ast})-J(\boldsymbol{x}^{adv}_{t})\right) \\
    & \leq \frac{D_{1}^2\sqrt{T}}{2\alpha } + \frac{{\mu D_{1}}^2}{2\alpha (1-\lambda)^2} +  2\alpha {D_{2}}^2 \sqrt{T}.
   \end{aligned}
\end{equation*}
Thus.
\begin{equation*}
\begin{aligned}
    & \frac{G}{T}\sum_{t=1}^{T}\left( J(\boldsymbol{x}^{\ast})-J(\boldsymbol{x}^{adv}_{t}) \right) \\
   &  \leq \frac{D_{1}^2}{2\alpha \sqrt{T}} + \frac{{\mu D_{1}}^2}{2\alpha (1-\lambda)^2 T} +  \frac{2\alpha {D_{2}}^2}{\sqrt{T}}.
   \end{aligned}
 \end{equation*}
 By concavity of $J(\boldsymbol{x})$, we obtain
 \begin{equation}\label{f}
   \begin{aligned}
   &  J(\boldsymbol{x}^{\ast})-J(\boldsymbol{\bar{x}}^{adv}_T) \\
   &  \leq G \left(\frac{D_{1}^2}{2\alpha \sqrt{T}} + \frac{{\mu D_{1}}^2}{2\alpha (1-\lambda)^2 T} +  \frac{2\alpha {D_{2}}^2}{\sqrt{T}}\right).
   \end{aligned}
 \end{equation}

This completes the proof of Lemma \ref{lem:Simple AdaMI}.\\

The original analysis of AdaGrad \cite{Duchi2010AdaptiveSM} focused on the online case, in which a data-dependent optimal regret bound of order $O(d \sqrt T)$ was obtained. Then, the rate of associated stochastic algorithms can be derived using a standard online-to-batch conversion. In \cite{mukkamala2017variants}, it has been indicated RMSProp contains AdaGrad as a special case for a particular choice of the weighting scheme. Moreover, a general analysis of RMSProp was given for both convex and strongly-convex functions. The detailed analysis \cite{mukkamala2017variants} implies that one can derive similar convergence rates for the adaptive variants of the predetermined step size methods without additional difficulties. Specifically, to extend Lemma \ref{lem:Simple AdaMI} to an adaptive setting, we first conduct the analysis for each coordinate of $\bm{x}$ like that in Lemma \ref{lem:Simple AdaMI}. Note that each $v_{t,i}$ in AdaMI is the EMA of the square of the $i$-th elements of the past momentums, which is different from that of the past gradients in Adam. However, this distinction does not affect the proof. In fact, as $\boldsymbol{g}_{t}$ is bounded (Lemma \ref{lem:usefullemma1}), $\displaystyle\frac{\alpha }{\sqrt{t}} \hat{V_t}^{-\frac{1}{2}} $ decreases generally on the order of $O(\frac{1}{\sqrt t})$ like that in AdaGrad, and we can get similar bounds of $P_1$, $P_2$ and $P_3$ by summing all the derived bounds from each coordinate. Thus, Theorem \ref{maintheorem1} can be proved like that in \cite{mukkamala2017variants, reddi2019convergence}.

\subsection{More Momentum-based Attack Experiments}

In order to verify the effectiveness of the proposed attack method more comprehensively, we also use Inc-v3, VGG16, ViT-S, Visformer-S, Swin-T as source models. The attack success rates are reported in Tab.\ref{tab:momentum_all}. As can be seen, our derived momentum-based adaptive attacks achieve the better transferability over the corresponding attacks.

\begin{table*}[htbp]\small
\captionsetup{font={normalsize}}
\caption{\normalsize{Transferability comparisons with different momentum-based attacks. * indicates the results on the white-box model. The best results are marked in bold.}}
\label{tab:momentum_all}
\centering
\resizebox{\textwidth}{!}{
\begin{tabular}{c|c|cccccccc}
\toprule
Model& Attack& Res34& Inc-v3& VGG16& Mob-v2& ViT-S& ConViT-B& Visformer-S& Swin-T \\
\midrule 
\multirow{16}{*}{Res34}& MI& 100.0$^*$& 36.8& 47.2& 45.8& 17.4& 12.3& 19.0& 25.9 \\
\multirow{16}{*}{ }& AdaMI (Ours)& 100.0$^*$& 38.4& 51.5& 49.2& 18.4& 13.3& 20.0& 28.0\\	
\multirow{16}{*}{ }&  NI& 100.0$^*$&38.1&48.6&47.0&17.8&12.8&19.6&27.4\\
\multirow{16}{*}{ }& AdaNI (Ours)& 100.0$^*$&39.6&51.6&50.7& 19.2&13.1&20.6&27.8\\
\multirow{16}{*}{ }& VMI& 100.0$^*$& 48.4& 61.1& 57.0& 25.0& 18.0& 28.1& 36.4\\
\multirow{16}{*}{ }& AdaVMI (Ours)& 100.0$^*$& 49.0& 61.4& 57.4& 25.6& 18.1& 28.5& 37.1\\
\multirow{16}{*}{ }& IE&100.0$^*$&40.2&52.6&50.3&19.6&13.3&20.7&29.1\\
\multirow{16}{*}{ }& AdaIE (Ours)& 100.0$^*$& 41.1& 53.5& 50.7& 19.6& 13.2& 20.9& 28.9\\
\multirow{16}{*}{ }& EMI& 100.0$^*$& 47.4& 62.5& 58.5& 22.1& 16.7& 24.5& 35.2\\
\multirow{16}{*}{ }& AdaEMI (Ours)& 100.0$^*$&47.9&62.8&59.7&21.9&15.9&25.8&35.4\\
\multirow{16}{*}{ }& MIG& 100.0$^*$& 52.2& 58.5& 53.7& 21.8& 16.0& 23.8& 32.1\\
\multirow{16}{*}{ }& AdaMIG (Ours)& 100.0$^*$& 52.5& 57.8& 54.4& 22.1& 16.7& 24.2& 32.3\\
 \multirow{16}{*}{ }& PGN& 100.0$^*$& 59.5& 71.7& 66.1& 32.3& 25.0& 36.7& 44.8\\
 \multirow{16}{*}{ }& AdaPGN (Ours)& 100.0$^*$& 59.9& 72.6& 67.6& 32.5& 25.5& 37.6& 45.4\\
\multirow{16}{*}{}& NCS&100.0$^*$& 62.8&80.2&51.5&35.6&49.5&52.2&34.9 \\
\multirow{16}{*}{}& AdaNCS (Ours)& 100.0$^*$ & \bf{64.5} & \bf{83.2} & \bf{53.9} & \bf{37.4} & \bf{52.2} & \bf{54.3} & \bf{36.5} \\ \hline
\multirow{16}{*}{Inc-v3}& MI& 24.9& 97.1$^*$& 29.3& 30.0& 13.1& 8.3& 13.2& 17.5 \\
\multirow{16}{*}{ }& AdaMI (Ours)& 30.3& 98.1$^*$& 34.5& 35.4& 15.4& 10.3& 16.1& 21.1\\
\multirow{16}{*}{ }&  NI& 29.4&97.3$^*$&30.7&32.8&13.8& 9.9&14.8&20.2\\
\multirow{16}{*}{ }& AdaNI (Ours)& 33.0& 97.5$^*$& 35.9& 36.9& 16.9& 10.9& 16.2& 22.0\\
\multirow{16}{*}{ }&  VMI & 33.8& 97.4$^*$& 34.5& 37.1& 16.0& 12.0& 17.9& 22.5\\
\multirow{16}{*}{ }& AdaVMI (Ours)& 37.6& 98.7$^*$& 40.3& 40.8& 18.3& 12.6& 19.8& 23.8\\
\multirow{16}{*}{ }& IE&  29.3&97.8$^*$&31.9&33.3&15.0&9.9&15.5&20.6\\
\multirow{16}{*}{ }& AdaIE (Ours)& 30.0& 98.0$^*$& 32.0&33.8 &14.7&10.4&15.7&20.5\\
\multirow{16}{*}{ }& EMI&38.3&99.3$^*$&39.6& 39.8& 17.8&12.7&19.5&25.2\\
\multirow{16}{*}{ }& AdaEMI (Ours)& 38.7& 99.6$^*$& 40.4& 42.4& 19.0& 12.8& 20.1& 24.1\\
\multirow{16}{*}{ }& MIG& 35.7& 98.7$^*$& 36.5& 38.6& 15.6& 10.6& 16.1& 21.9\\
\multirow{16}{*}{ }& AdaMIG (Ours)& 35.3& 98.8$^*$& 36.7& 37.0& 15.1& 10.7& 16.5& 22.1\\
\multirow{16}{*}{ }& PGN& 47.2& 99.6$^*$& 46.3& 47.0& 23.5& 19.2& 23.8& 30.8\\
\multirow{16}{*}{ }& AdaPGN (Ours)& 47.5& 99.6$^*$& 47.0& 46.1& 22.4& 18.6& 25.7& 31.4\\ 
\multirow{16}{*}{ }& NCS&54.4 & 98.2$^*$ & 53.2 & 52.6 & 27.6 & 23.4 & 29.9 & 35.8 \\
\multirow{16}{*}{ }& AdaNCS (Ours)&\bf{58.2} & 98.5$^*$ & \bf{54.2} & \bf{55.0} & \bf{29.4} & \bf{24.2} & \bf{32.0} & \bf{39.1}\\
\hline
\multirow{16}{*}{VGG16}& MI& 35.2& 29.2& 99.3$^*$& 41.5& 13.9& 8.8& 16.1& 23.3\\
\multirow{16}{*}{ }& AdaMI (Ours)& 37.4& 30.0& 99.6$^*$& 47.9& 14.8& 9.0& 18.4& 25.4\\
\multirow{16}{*}{ }&  NI& 35.1&30.8&99.4$^*$&45.7&13.3& 8.6&16.8&24.1\\
\multirow{16}{*}{ }& AdaNI (Ours)& 37.4& 30.1& 99.7$^*$& 48.1& 14.7& 9.0& 17.8& 26.2\\
\multirow{16}{*}{ }&  VMI & 48.7& 39.4& 99.8$^*$& 54.3& 18.9& 12.9& 24.1& 33.1\\
\multirow{16}{*}{ }& AdaVMI (Ours)& 49.8& 39.1& 99.8$^*$& 56.2& 19.0& 13.1& 24.8& 33.3\\
\multirow{16}{*}{ }& IE &  39.2&31.4&99.7$^*$&48.4&15.0&9.1&18.7&26.0\\
\multirow{16}{*}{ }& AdaIE (Ours)&39.7&30.9& 99.7$^*$&46.6&15.0&9.0&18.7&26.4\\
\multirow{16}{*}{ }& EMI &46.2&35.1&99.9$^*$& 55.2& 16.4&9.8&21.7&29.9\\
\multirow{16}{*}{ }& AdaEMI (Ours) & 45.9& 35.1& 99.9$^*$& 55.9& 16.3& 10.2& 22.0& 31.2\\
\multirow{16}{*}{ }& MIG & 48.3& 40.9& 99.9$^*$& 53.6& 17.4& 11.1& 21.2& 29.4\\
\multirow{16}{*}{ }& AdaMIG (Ours)& 47.2& 41.6& 99.9$^*$& 53.7& 16.6& 11.4& 21.4& 28.5\\
\multirow{16}{*}{ }& PGN& 59.8& 48.3& 99.9$^*$& 66.2& 22.9& 16.1& 30.7& 40.3\\
\multirow{16}{*}{ }& AdaPGN (Ours)& 60.2& 48.8& 99.9$^*$& 65.5 & 23.7& 17.0& 31.0& 39.9\\
\multirow{2}{*}{ }& NCS &65.4 & 51.3 & 99.8$^*$ & 69.9 & 26.8 & 19.4 & 36.1 & 45.0\\
\multirow{2}{*}{ }& AdaNCS (Ours) & \bf{68.5} & \bf{53.5} & 99.8$^*$ & \bf{72.6} & \bf{28.2} & \bf{20.7} & \bf{37.4} & \bf{46.5}\\
\hline
\end{tabular}}
\end{table*}

\begin{table*}[htbp]\small
\ContinuedFloat
\centering
\resizebox{\textwidth}{!}{
\begin{tabular}{c|c|cccccccc}
\toprule
Model& Attack  &  Res34& Inc-v3& VGG16& Mob-v2& ViT-S& ConViT-B& Visformer-S& Swin-T \\
\midrule 
\multirow{16}{*}{ViT-S }& MI& 29.9& 30.6& 32.1& 35.9& 99.6$^*$& 41.4& 24.0& 38.3\\
\multirow{16}{*}{ }& AdaMI (Ours)& 30.5& 31.2& 35.4& 37.6& 99.8$^*$& 42.8& 25.6& 38.8\\
\multirow{16}{*}{ }& NI& 30.2& 31.5& 35.5& 39.1& 99.4$^*$& 41.6& 26.2& 39.0\\
\multirow{16}{*}{ }& AdaNI (Ours) & 31.8& 33.9& 37.9& 41.0& 99.8$^*$& 42.0& 27.1& 40.6\\
\multirow{16}{*}{ }& VMI & 34.7& 34.7& 37.6& 40.8& 99.6$^*$& 50.8& 31.5& 43.0\\
\multirow{16}{*}{ }& AdaVMI (Ours) & 34.7& 35.2& 38.6& 41.4& 99.9$^*$& 49.9& 31.6& 44.1\\
\multirow{16}{*}{ }& IE & 32.7& 32.3& 36.2& 39.8& 99.9$^*$& 47.5& 29.5& 41.6\\
\multirow{16}{*}{ }& AdaIE (Ours) & 33.1& 32.8& 35.6& 40.4& 99.9$^*$& 47.4& 28.4& 41.8\\ 
\multirow{16}{*}{}& EMI & 38.1& 39.1& 44.8& 46.8& 100.0$^*$& 55.6& 32.8&49.7
\\
\multirow{16}{*}{ }& AdaEMI (Ours) & 39.3& 38.5& 43.8& 47.0& 100.0$^*$& 55.8& 33.6&50.1
\\
\multirow{16}{*}{ }& MIG & 36.6& 38.4& 38.1& 42.7& 98.3$^*$& 50.1& 30.3&44.7 
\\
\multirow{16}{*}{ }& AdaMIG (Ours)& 37.5& 38.4& 37.8& 43.4& 98.6$^*$& 49.9& 30.7&44.9
\\
\multirow{16}{*}{ }& PGN& 44.8& 43.5& 47.9& 50.1& 100.0$^*$& 64.0& 40.8&55.4
\\
\multirow{16}{*}{ }& AdaPGN (Ours)& 44.8& 45.0& 48.6& 50.5 & 100.0$^*$& 64.5& 40.7&55.5
\\
\multirow{16}{*}{ }& NCS& 47.5 & 46.0 & 51.4 & 53.5 & 99.4$^*$ & 66.7 & 45.8 &58.5
\\
\multirow{16}{*}{ }& AdaNCS (Ours)& \bf{50.9} & \bf{49.0} & \bf{53.6} & \bf{55.0} & 99.7$^*$ & \bf{ 70.8} & \bf{48.1} &\bf{62.0}\\ 
\hline
\multirow{16}{*}{Visformer-S}& MI& 32.9& 32.0& 41.7& 41.5& 23.0& 22.4& 87.4$^*$& 40.5\\
\multirow{16}{*}{ }& AdaMI (Ours)& 34.6& 33.1& 43.4& 44.0& 22.5& 22.3& 93.4$^*$& 42.4\\
\multirow{16}{*}{ }& NI& 31.7& 31.8& 40.6& 42.1& 21.4& 20.0& 90.0$^*$& 41.7\\
\multirow{16}{*}{ }& AdaNI (Ours)& 34.0& 34.0& 44.0& 44.5& 22.2& 22.2& 92.8$^*$& 42.7\\
\multirow{16}{*}{ }& VMI& 43.2& 42.9& 51.7& 51.0& 34.7& 35.8& 90.1$^*$& 52.8\\
 \multirow{16}{*}{ }& AdaVMI (Ours)& 45.1& 43.1& 54.3& 52.4& 34.9& 34.9& 91.3$^*$& 53.3\\
 \multirow{16}{*}{ }& IE& 34.5& 34.2& 44.8& 45.5& 25.3& 24.6& 93.2$^*$& 45.6\\
 \multirow{16}{*}{ }& AdaIE (Ours)& 34.7& 34.4& 43.9& 46.1& 26.1& 24.9& 92.8$^*$& 46.2\\
 \multirow{16}{*}{ }& EMI& 46.1& 41.8& 55.3& 54.8& 33.8& 32.4& 96.5$^*$& 55.7\\
 \multirow{16}{*}{ }& AdaEMI (Ours)& 45.9& 43.0& 55.3& 55.6& 33.3& 32.4& 96.5$^*$ & 56.1\\
 \multirow{16}{*}{ }& MIG& 46.0& 44.9& 52.9& 52.7& 37.9& 39.2& 88.5$^*$& 55.3\\
 \multirow{16}{*}{ }& AdaMIG (Ours)& 46.0& 45.9& 52.8& 51.8& 38.1& 39.5& 88.6$^*$& 56.1\\
 \multirow{16}{*}{ }& PGN& 57.1& 55.7& 61.0& 59.9& 50.2& 53.8& 88.0$^*$& 65.2\\
 \multirow{16}{*}{ }& AdaPGN (Ours)& 57.9& 55.8& 61.8& 60.2& 49.9& 53.5& 87.9$^*$& 65.4\\ 
 \multirow{16}{*}{ }&NCS&59.4 & 56.1 & 62.6 & 63.6 & 51.2 & 52.7 & 89.9$^*$ & 67.1\\
\multirow{16}{*}{ }&AdaNCS (Ours)&\bf{63.7} & \bf{60.6} & \bf{69.4} & \bf{68.4} & \bf{55.7} & \bf{59.0} & 93.7$^*$ & \bf{72.1} \\ 
 \hline
\multirow{16}{*}{Swin-T}& MI& 23.2& 23.7& 28.9& 33.7& 17.0& 14.1& 19.8& 84.0$^*$\\
\multirow{16}{*}{ }& AdaMI (Ours)& 22.9& 22.6& 31.5& 34.8& 16.6& 13.8& 21.1& 84.2$^*$\\
\multirow{16}{*}{ }& NI& 21.7& 23.2& 29.0& 34.2& 15.2& 13.4& 20.0& 81.6$^*$\\
\multirow{16}{*}{ }& AdaNI (Ours)& 23.0& 23.3& 30.8& 36.3& 16.4& 14.7& 20.9& 82.9$^*$\\
\multirow{16}{*}{ }& VMI& 34.7& 34.3& 40.7& 45.5& 30.2& 29.3& 36.7& 90.7$^*$\\
\multirow{16}{*}{ }& AdaVMI (Ours)& 34.2& 35.4& 41.3& 46.2& 30.5& 29.6& 37.5& 92.7$^*$\\
\multirow{16}{*}{ }& IE& 23.2& 22.9& 29.8& 35.2& 17.0& 13.9& 21.0& 88.1$^*$\\
\multirow{16}{*}{ }& AdaIE (Ours)& 23.5& 23.3& 29.8& 34.7& 17.3& 13.9& 21.1& 89.1$^*$\\
\multirow{16}{*}{ }& EMI& 28.8& 28.0& 37.5& 40.7& 22.4& 20.0& 28.7& 93.4$^*$\\
\multirow{16}{*}{ }& AdaEMI (Ours)& 29.6& 28.8& 39.6& 43.4& 22.5& 19.5& 28.7& 93.3$^*$\\
\multirow{16}{*}{ }& MIG& 30.7& 30.4& 37.0& 40.3& 25.7& 22.2& 29.3& 87.0$^*$\\
\multirow{16}{*}{ }& AdaMIG (Ours)& 30.9& 30.4& 36.6& 40.8& 25.7& 21.9& 30.5& 87.1$^*$\\ 
\multirow{16}{*}{ }& PGN& 57.1& 54.1& 61.8& 65.1& 52.3& 56.8& 62.5& 95.0$^*$\\ 
\multirow{16}{*}{ }& AdaPGN (Ours)& 58.2& 55.4& 62.0& 65.7& 54.4& 56.6& 62.8 & 94.7$^*$\\
\multirow{16}{*}{ }& NCS & 58.2 & 53.7 & 63.7 & 65.1 & 51.7 & 55.4 & 63.3 & 96.2$^*$ \\
\multirow{16}{*}{ }& AdaNCS (Ours) & \bf{60.3} & \bf{56.8} & \bf{67.1} & \bf{69.6} & \bf{54.7} & \bf{58.6} & \bf{66.8} & 98.2$^*$\\
\hline
\end{tabular}}
\end{table*}

\end{appendix}

\end{document}